\documentclass{article}

\usepackage{microtype}
\usepackage{graphicx}
\usepackage{booktabs} 

\usepackage{amsmath}
\usepackage{amssymb}
\usepackage{mathtools}
\usepackage{relsize}

\usepackage{multirow}
\usepackage{adjustbox}
\usepackage{array}
\usepackage{arydshln}
\usepackage{tabularx}
\usepackage{multirow}
\usepackage{lscape}
\usepackage{rotating}

\usepackage{subcaption}
\usepackage{afterpage}
\usepackage{comment}

\usepackage{hyperref}
\usepackage{kotex}

\usepackage[accepted]{icml2020}

\icmltitlerunning{Submission and Formatting Instructions for ICML 2020}

\begin{document}

\twocolumn[
\icmltitle{Class-Imbalanced Semi-Supervised Learning}



\icmlsetsymbol{equal}{*}

\begin{icmlauthorlist}
\icmlauthor{Minsung Hyun}{equal,snu,skh}
\icmlauthor{Jisoo Jeong}{equal,snu}
\icmlauthor{Nojun Kwak}{snu}
\end{icmlauthorlist}

\icmlaffiliation{snu}{Seoul National University}
\icmlaffiliation{skh}{SK hynix}

\icmlcorrespondingauthor{Nojun Kwak}{nojunk@snu.ac.kr}

\icmlkeywords{Machine Learning, ICML}

\vskip 0.3in
]



\printAffiliationsAndNotice{\icmlEqualContribution} 

\begin{abstract}

Semi-Supervised Learning (SSL) has achieved great success in overcoming the difficulties of labeling and making full use of unlabeled data.
However, SSL has a limited assumption that the numbers of samples in different classes are balanced, and many SSL algorithms show lower performance for the datasets with the imbalanced class distribution.
In this paper, we introduce a task of class-imbalanced semi-supervised learning (CISSL), which refers to semi-supervised learning with class-imbalanced data. In doing so, we consider class imbalance in both labeled and unlabeled sets.
First, we analyze existing SSL methods in imbalanced environments and examine how the class imbalance affects SSL methods.
Then we propose Suppressed Consistency Loss (SCL), a regularization method robust to class imbalance.
Our method shows better performance than the conventional methods in the CISSL environment.
In particular, the more severe the class imbalance and the smaller the size of the labeled data, the better our method performs.
\end{abstract}

\section{Introduction}
\label{sec:introduction}

A large dataset with well-refined annotations is essential to the success of deep learning and every time we encounter a new problem, we should annotate the whole dataset, which costs a lot of time and effort~\cite{russakovsky2015best,bearman2016s}.
To alleviate this annotation burden, many researchers have studied semi-supervised learning (SSL) that improves the performance of models by utilizing the information contained in unlabeled data~\cite{chapelle2009semi,verma2019interpolation, berthelot2019mixmatch}.

However, SSL has a couple of main assumptions and shows excellent performance only in these limited settings.
The first assumption is that unlabeled data is in-distribution, i.e., the class types of unlabeled data are the same as those of labeled data~\cite{oliver2018realistic}.
The second is the assumption of balanced class distribution, which assumes that each class has almost the same number of samples
~\cite{li2011semi, stanescu2014semi}.
In this paper, we performed a study dealing with the second assumption. 

\begin{figure*}[!t]
    \centering
    \begin{subfigure}{0.23\linewidth}
        \centering
        \includegraphics[width=4cm, height=3.5cm]{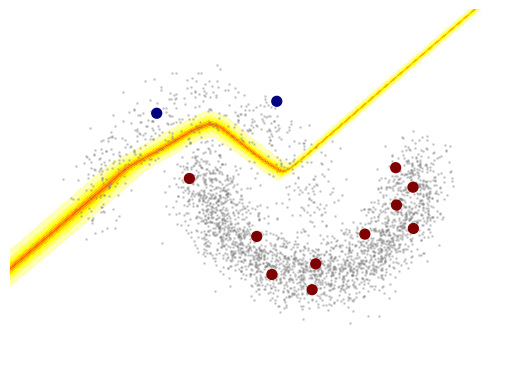}
        \caption{Two moons Supervised}
        \label{fig:toy_twomoons_sup}
    \end{subfigure}
    \begin{subfigure}{0.23\linewidth}
        \centering
        \includegraphics[width=4cm, height=3.5cm]{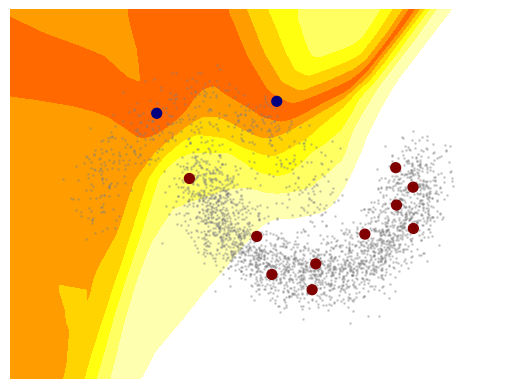}
        \caption{Two moons $\Pi$ model}
        \label{fig:toy_twomoons_pi}
    \end{subfigure}
    \begin{subfigure}{0.23\linewidth}
        \centering
        \includegraphics[width=4cm, height=3.5cm]{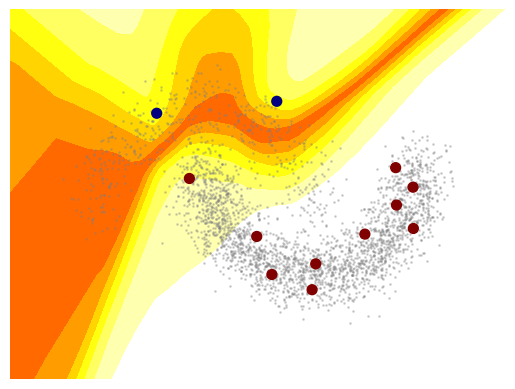}
        \caption{Two moons Mean Teacher}
        \label{fig:toy_twomoons_mt}
    \end{subfigure}
    \begin{subfigure}{0.27\linewidth}
        \centering
        \includegraphics[width=4.7cm, height=3.5cm]{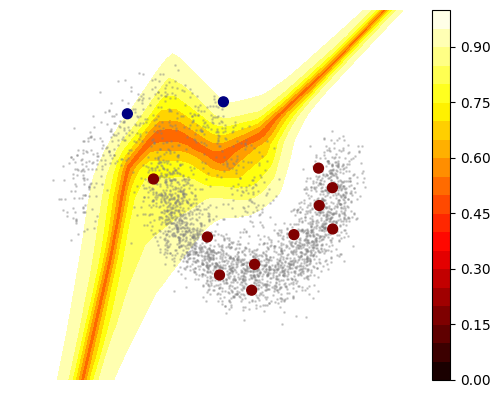}
        \caption{Two moons SCL (ours)}
        \label{fig:toy_twomoons_scl}
    \end{subfigure}
    
    \begin{subfigure}{0.23\linewidth}
        \centering
        \includegraphics[width=4cm, height=3.5cm]{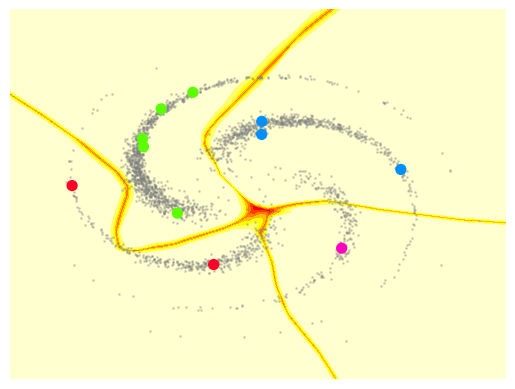}
        \caption{Four spins Supervised}
        \label{fig:toy_fourspins_sup}
    \end{subfigure}
    \begin{subfigure}{0.23\linewidth}
        \centering
        \includegraphics[width=4cm, height=3.5cm]{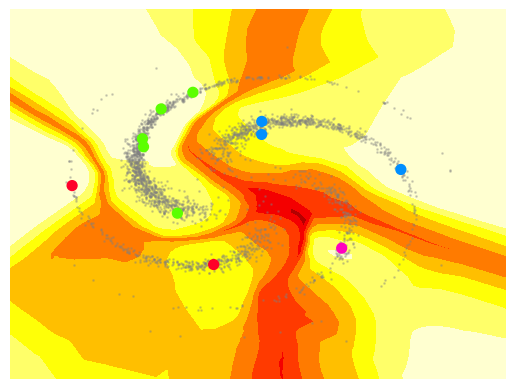}
        \caption{Four spins $\Pi$ model}
        \label{fig:toy_fourspins_pi}
    \end{subfigure}
    \begin{subfigure}{0.23\linewidth}
        \centering
        \includegraphics[width=4cm, height=3.5cm]{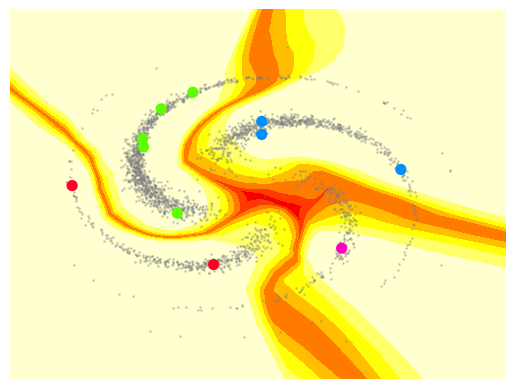}
        \caption{Four spins Mean Teacher}
        \label{fig:toy_fourspins_mt}
    \end{subfigure}
    \begin{subfigure}{0.27\linewidth}
        \centering
        \includegraphics[width=4.7cm, height=3.5cm]{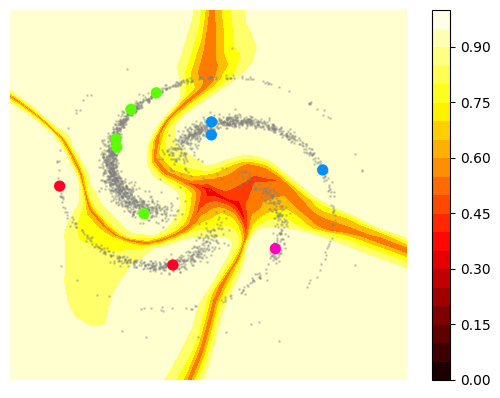}
        \caption{Four spins SCL (ours)}
        \label{fig:toy_fourspins_scl}
    \end{subfigure}
\vspace{-2mm}
\caption{Toy examples: We experimented on \textit{Two moons} and \textit{Four spins} datasets in CISSL settings for four algorithms (Supervised learning, $\Pi$ model~\cite{laine2016temporal}, Mean Teacher~\cite{tarvainen2017mean} and SCL (ours)). The color represents the probability of of the class with the highest confidence.}
\label{fig:toy}
\vspace{-2mm}
\end{figure*}

The class distribution of data, in reality, is not refined and is known to have long tails~\cite{kendall1946advanced}.
However, many researches have developed models based on well-refined balanced data such as CIFAR~\cite{krizhevsky2009learning}, SVHN~\cite{netzer2011reading}, and ImageNet ILSCVRC 2012~\cite{deng2009imagenet}.
Training the model with imbalanced datasets causes performance degradation.
Class imbalanced learning (CIL) is a way to solve such class imbalance and proposes various methods in the level of data, algorithm, and their hybrids~\cite{krawczyk2016learning, johnson2019survey}.
However, to our best knowledge, the studies on CIL have relied entirely on labeled datasets for training and have not considered the use of unlabeled data.

In this paper, we define a task, \textit{class-imbalanced semi-supervised learning} (CISSL), and propose a suitable algorithm for it.
By assuming class imbalance in both labeled and unlabeled data, CISSL relaxes the assumption of balanced class distribution in SSL. Also, it can be considered as a task of adding unlabeled data to CIL.

We analyzed the existing SSL methods in the CISSL setting through toy examples.
First, we found that the class imbalance in the CISSL disrupts the learning of the existing SSL methods based on the \textit{`cluster assumption'} which asserts that each class has its own cluster in the latent space~\cite{chapelle2009semi}.
According to this assumption, the decision boundary traverses the low-density area of the latent space.
With the class imbalance, however, the decision boundary may be incorrectly formed and passes through the high-density area of the minor class, which results in degradation of the SSL methods.

In Fig.\ref{fig:toy_twomoons_pi}, \ref{fig:toy_fourspins_pi}, we can see that each decision boundary is skewed toward the minority class in the $\Pi$ model~\cite{laine2016temporal}, a representative algorithm of consistency-regularization-based SSL, compared to that of supervised learning (Fig.\ref{fig:toy_twomoons_sup}, \ref{fig:toy_fourspins_sup}).

Second, we examined that the Mean Teacher (MT)~\cite{tarvainen2017mean} is more robust than $\Pi$ model in CISSL settings.
In Fig.\ref{fig:toy_twomoons_mt}, \ref{fig:toy_fourspins_mt}, even though there is a class imbalance, MT maintains a relatively stable decision boundary.
We show later that MT is more stable because it uses a conservative target for consistency regularization.

Based on these observations, we propose a regularization method using \textit{`suppressed consistency loss'} (SCL), for better performance in the CISSL settings.
SCL prohibits the decision boundary in a minor class region from being smoothed too much in the wrong direction as shown in Fig.\ref{fig:toy_twomoons_scl}, \ref{fig:toy_fourspins_scl}.
In Section \ref{sec:scl}, we will discuss the role of SCL in more detail.

We also proposed standard experimental settings in the CISSL.
We followed the SSL experiment settings, but to be more realistic, we considered class imbalance in both labeled and unlabeled data.
In this setting, we compared existing SSL and CIL methods to ours and found that our method with SCL shows better performance than others.
Furthermore, we applied SCL to the object detection problem and improved performance in the existing SSL algorithm for object detection.

Our main contributions can be summarized as follows:\\
$\bullet$ We defined a task of imbalanced semi-supervised learning, reflecting a more realistic situation, and suggested standard experimental settings. \\
$\bullet$ We analyzed how the existing SSL methods work in CISSL settings through mathematical and experimental results. \\
$\bullet$ We proposed Suppressed Consistency Loss that works robustly for problems with class imbalance, and experimentally show that our method improves performance.

\section{Related Work}
\label{sec:related_works}

\subsection{Semi-Supervised Learning}
\label{sec:ssl}
Semi-supervised learning is a learning method that tries to improve the performance of supervised learning, which is based only on labeled data ($\mathcal{D_{L}}$), by additional usage of unlabeled data ($\mathcal{D_{U}}$).
SSL approaches include methods based on self-training and generative models~\cite{lee2013pseudo, zhai2019s4l, goodfellow2014generative, radford2015unsupervised, dumoulin2016adversarially, lecouat2018manifold}.
In addition, consistency regularization has shown good performance in semi-supervised learning, which pushes the decision boundary to low-density areas using unlabeled data~\cite{bachman2014learning, sajjadi2016regularization, laine2016temporal, verma2019interpolation}.
The objective function $\mathcal{J}$ is composed of supervised loss, $L_{sup}$, for $\mathcal{D_{L}}$ and consistency regularization loss, $L_{con}$, for $\mathcal{D_{U}}$.
As a typical semi-supervised learning method \cite{laine2016temporal, oliver2018realistic}, ramp-up scheduling function $w(t)$ is used for stable training:
\begin{gather}
    {\mathcal{J}} = L_{sup} + w(t) \cdot L_{con} 
    \label{eq:objective_function} \\
    L_{con}(X) = d(f_\theta(X+\epsilon), f_{\theta_{tg}}(X+\epsilon')), 
    \label{eq:consis_loss}
\end{gather}
where $d$ is a distance metric such as $L_2$ distance or KL-divergence, $\epsilon$ and $\epsilon'$ are perturbations to input data, and $\theta$ and $\theta_{tg}$ are the parameters of the model and target model, respectively. For $C$-class classification problem, $f_\theta(X) \in \mathbb{R}_+^C$ is the output logit (class probability) for the input $X$.
$\Pi$ model~\cite{laine2016temporal} and Mean Teacher (MT)~\cite{tarvainen2017mean} are the representative algorithms using consistency regularization.
The $\Pi$ model uses $\theta$ as $\theta_{tg}$ and MT updates $\theta_{tg}$ with EMA (exponential moving average) as follows:
\begin{equation}
    \theta_{tg} \leftarrow \gamma \theta_{tg} + (1 - \gamma) \theta.
    \label{eq:mt_target}
\end{equation}
From (\ref{eq:mt_target}), MT can be considered as a temporal ensemble model in the parameter space.

Above this, there are some methods that optimize the direction of perturbation~\cite{miyato2018virtual}, regularize through graphs of minibatch samples~\cite{luo2018smooth} and perturb inputs with mixup~\cite{zhang2017mixup, verma2019interpolation}.
In addition, the consistency-based semi-supervised learning for object detection (CSD) is an algorithm that applies SSL to object detection by devising classification and localization consistency~\cite{jeong2019consistency}.

\subsection{Class Imbalanced Learning}
\label{sec:cil}

Class imbalanced learning is a way to alleviate the performance degradation due to class imbalance.
\citet{buda2018systematic} defined the class imbalance factor $\rho$ as the ratio between the numbers of samples of the most frequent and the least frequent classes.
And we call each class as major class and minor class.

So far, there have been various researches to solve class imbalance problems~\cite{johnson2019survey}.
Data-level methods approach the problem by over-sampling minor classes or under-sampling major classes~\cite{masko2015impact, lee2016plankton, pouyanfar2018dynamic, buda2018systematic}.
These methods take a long time in model training due to re-sampling.
Algorithm-level methods re-weight the loss or propose a new loss without touching the sampling scheme~\cite{wang2016training, lin2017focal, wang2018predicting, khan2017cost, zhang2016training, wang2017learning, cui2019class, cao2019learning}.
Algorithm-level methods can be easily applied without affecting training time.
There are also hybrids of both methods~\cite{huang2016learning, ando2017deep, dong2019imbalanced}.

In this paper, we applied three algorithm-level methods to the CISSL environment and compared their performance to cross-entropy loss (CE): \\
(\romannumeral 1) Normalized weights, which weight a loss inversely proportional to the class frequency (IN)~\cite{cao2019learning}.\\
(\romannumeral 2) Focal loss which modulates by putting fewer weights on samples that the model is easy to classify~\cite{lin2017focal}.
\\
(\romannumeral 3) Class-balanced loss which re-weights the loss in inverse proportion to the effective number of classes (CB)~\cite{cui2019class}.

\begin{table}[!t]
    \centering
    \renewcommand{\tabcolsep}{0.7mm}
    \caption{Mean and standard deviation of validation error rates~(\%) for all, major, and minor classes in toy examples. We conducted 5 runs with different random seeds for class imbalance distribution.
    }
    \label{tab:toy}
    \vskip -0.15in
    \begin{center}
    \begin{small}
    \begin{sc}
    \adjustbox{max width=\linewidth}{
    \begin{tabularx}{1.6\linewidth}{lc|r|r|r|r}
        \toprule
         (\%) & Class Type & \multicolumn{1}{c|}{Supervised} & \multicolumn{1}{c|}{$\Pi$ model} & \multicolumn{1}{c|}{Mean Teacher} & \multicolumn{1}{c}{MT+SCL (Ours)}\\
         \midrule
 & All & 25.06 $\pm$ 12.43 & 41.57 $\pm$ 8.82 & 34.99 $\pm$ 9.98 & \textcolor{red}{\bf{24.39 $\pm$ 15.14}} \\
Twomoons & Major & 0.95 $\pm$ 1.24 & \textcolor{red}{\bf{0.00 $\pm$ 0.00}} & 0.01 $\pm$ 0.03 & 0.06 $\pm$ 0.07 \\
 & Minor & 49.17 $\pm$ 24.74 & 83.14 $\pm$ 17.64 & 69.96 $\pm$ 19.98 & \textcolor{red}{\bf{48.01 $\pm$ 31.04}} \\
 \midrule
 & All & 19.70 $\pm$ 6.70 & 17.79 $\pm$ 8.39 & 14.99 $\pm$ 8.46 & \textcolor{red}{\bf{10.91 $\pm$ 8.94}} \\
Fourspins & Major & 7.83 $\pm$ 5.43 & \textcolor{red}{\bf{4.75 $\pm$ 3.74}} & 4.76 $\pm$ 3.30 & 6.28 $\pm$ 3.26 \\
 & Minor & 49.39 $\pm$ 25.61 & 52.68 $\pm$ 31.17 & 43.29 $\pm$ 31.53 & \textcolor{red}{\bf{27.68 $\pm$ 36.48}} \\

 \bottomrule
    \end{tabularx}}
    \end{sc}
    \end{small}
    \end{center}
    \vskip -0.3in
\end{table}

\section{Analysis of SSL under Class Imbalance}
\label{sec:analysis}

In this section, we look into the topography of the decision boundary to see how the SSL algorithms work in the class-imbalanced environment.
First, we compare supervised learning with SSL's representative algorithms, $\Pi$ model \cite{laine2016temporal} and Mean Teacher \cite{tarvainen2017mean} via toy examples.
And we analyze why MT performs better in CISSL through a mathematical approach.

\begin{figure*}[!ht]
\setkeys{Gin}{width=\linewidth, keepaspectratio}
\begin{minipage}{0.59\textwidth}
    \centering
    \includegraphics[width=10cm]{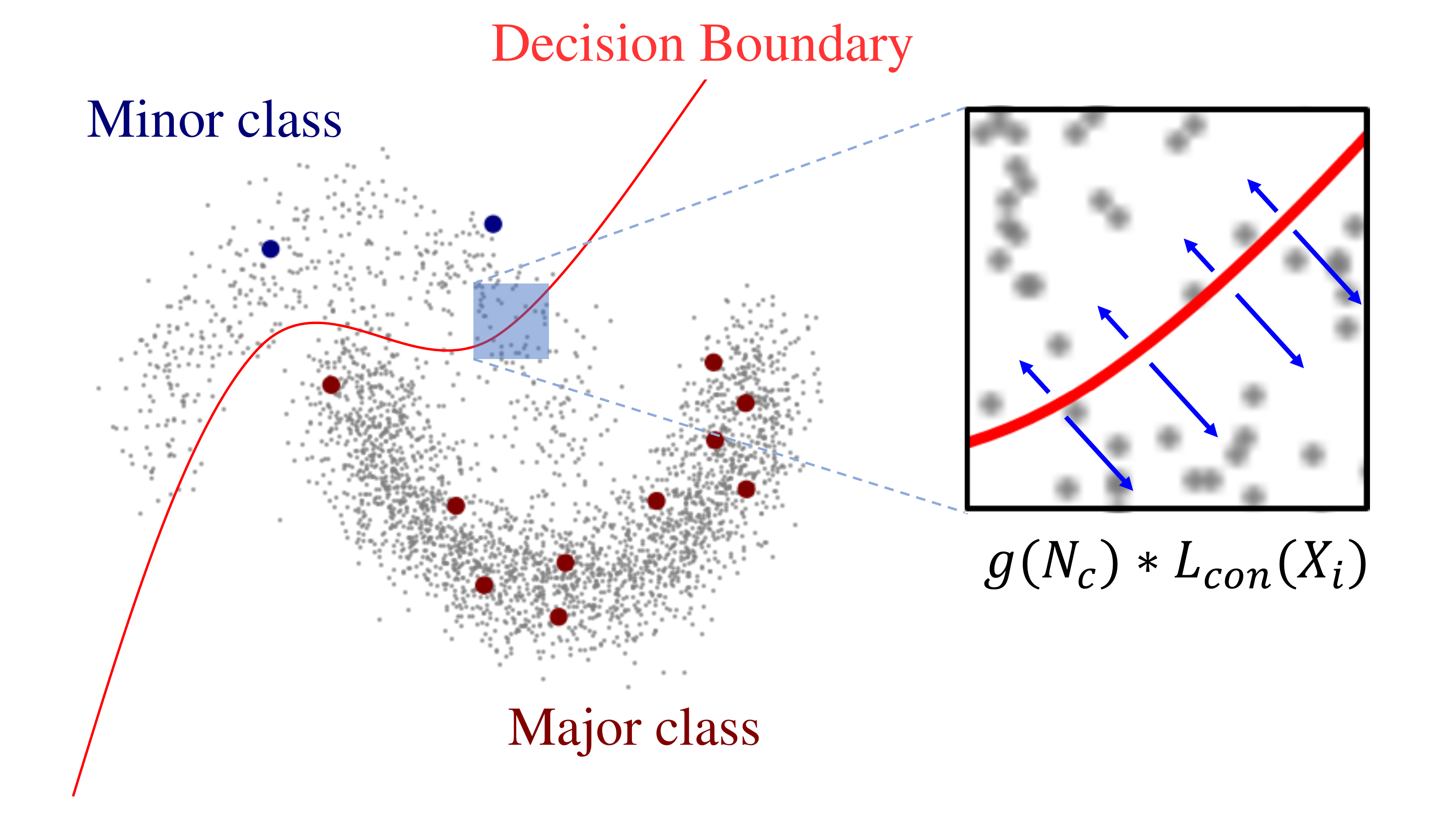}
    \caption{Suppressed Consistency Loss (SCL). Due to the imbalance in data, decision boundary tends to skew into the areas of minor class with consistency regularization. SCL inversely weights consistency loss to the number of class samples and pushes the decision boundary against low-density areas.}
    \label{fig:algorithm}

\end{minipage}\hfill
\begin{minipage}{0.39\textwidth}
    \centering
    \begin{subfigure}{0.45\linewidth}
        \includegraphics[width=3cm]{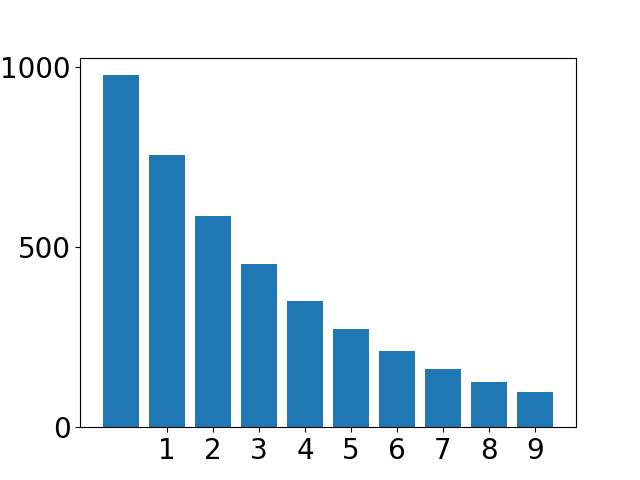}
        \caption{Labeled dataset}
    \end{subfigure}
    \begin{subfigure}{0.45\linewidth}
        \includegraphics[width=3cm]{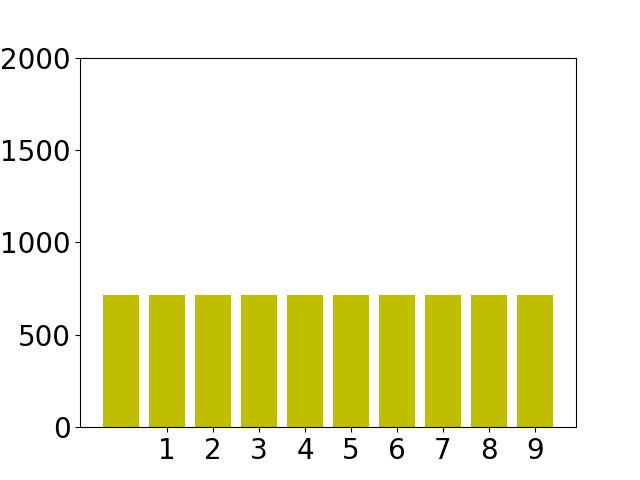}
        \caption{Uniform}
    \end{subfigure}
    
    \begin{subfigure}{0.45\linewidth}
        \includegraphics[width=3cm]{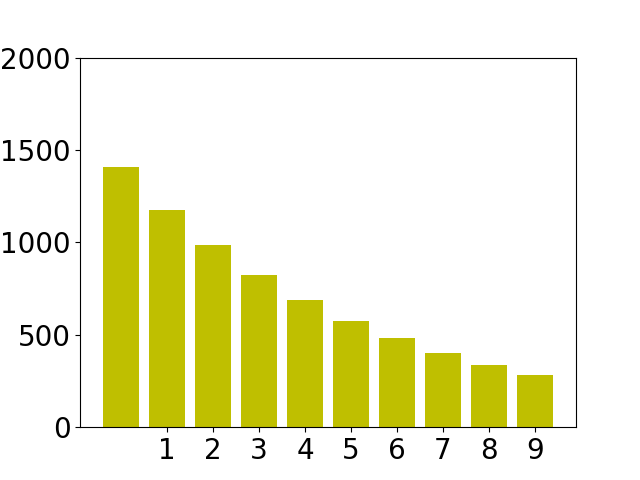}
        \caption{Half}
    \end{subfigure}
    \begin{subfigure}{0.45\linewidth}
        \includegraphics[width=3cm]{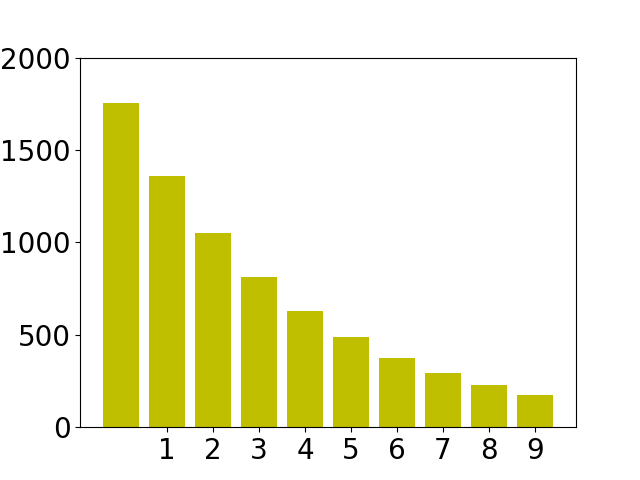}
        \caption{Same}
    \end{subfigure}
    \caption{Types of unlabeled data imbalance. (a) Imbalance of labeled data. (b) \textit{Uniform} case: The number of samples in all classes are the same ($\rho_u$ =1). (c) \textit{Half} case: The imbalance factor of the unlabeled data is half of the labeled data. ($\rho_u = \rho_l/2)$ (d) \textit{Same} case: Same labeled and unlabeled imbalance factor ($\rho_u = \rho_l)$.}
    \label{fig:type}
\end{minipage}
\vspace{-5mm}
\end{figure*} 

\subsection{Toy examples}
\label{sec:toy}

We trained each algorithm by 5,000 iterations on \textit{two moons} and \textit{four spins} datasets with an imbalance factor of 5 for each labeled and unlabeled data.
\footnote{The number of samples of class \textit{c} is set to \\ $N_c = N_{max} \times \rho^{-\frac{R_c - 1}{C - 1}}$. The \textit{Rank}, $R_c$, of the major class is 1.\\ $\rho$ and $C$ are the imbalance ratio and the number of classes.}
Fig.\ref{fig:toy} represents the probability of the class with the highest confidence at each location.
The region with relatively low probability, closer to the dark red color, is the decision boundary in the figure.

In Fig.\ref{fig:toy_twomoons_sup}, \ref{fig:toy_fourspins_sup}, the decision boundary of the supervised learning is very steep.
And there are very high confidence areas far away from the decision boundary.
With the SSL methods, unlabeled data smooth the decision boundary through consistency regularization~\cite{chapelle2009semi}.
In particular, the decision boundary smoothing is larger in the minor class area.
Also, we found that the learning patterns of the $\Pi$ model and MT are different.
Table.\ref{tab:toy} shows the validation error rates for toy examples. 
We found that performance degradation is evident in the minor class.
MT shows relatively better performance than $\Pi$ model, although it shows inferior performance than the supervised learning in \textit{two moons}.
Our method which applies SCL to MT achieves the best performance in both \textit{two moons} and \textit{four spins} datasets.

\subsection{$\Pi$ Model vs. Mean Teacher }
\label{sec:mathapproach}
We analyze the results of Section.\ref{sec:toy} in this part.
When the consistency regularization is applied to supervised learning in Fig.\ref{fig:toy_twomoons_sup}, \ref{fig:toy_fourspins_sup}, compared to the samples far away from the boundary, the influence of the samples around the decision boundary is considerable, because the model output does not change even if small perturbation is added to the model input in the region far from the decision boundary from (\ref{eq:consis_loss}).
As a result, consistency regularization smooths the decision boundary, as shown in Fig.\ref{fig:toy_twomoons_pi}, \ref{fig:toy_fourspins_pi}.

According to the \textit{cluster assumption}~\cite{chapelle2009semi}, the decision boundary lies in the low-density area and far from the high-density area.
However, in a problem with severe class imbalance, the decision boundary may penetrate a globally sparse but relatively high-density area of a minor class as shown in the blue square in Fig.\ref{fig:algorithm}.
By consistency regularization, decision boundary smoothing occurs in this area, and many samples in the minor class are misclassified.

Therefore, conventional consistency regularization-based methods are generally expected to degrade the performance for the minor class.
But we found that the severity of this phenomenon differs depending on the SSL algorithm.
In Table.\ref{tab:toy}, MT consistently performed better than $\Pi$ model, especially for the minor class.

First, we analyzed the behavior of MT in CISSL with the simple SGD optimizer.
Consider the model parameter $\theta$, the learning rate $\alpha$, and the objective function $\mathcal{J}$, then the update rule of SGD optimizer is:
\begin{equation}
    \theta \leftarrow \theta - \alpha \nabla \mathcal{J}(\theta).
    \label{eq:theta}
\end{equation}
For a EMA decay factor of MT, $\gamma \in (0, 1]$, the current ($\theta$) and the target ($\theta'$) model parameters at the $t$-th iteration are
\begin{equation}
    \theta_t = \theta_0 - \alpha \sum_{k=0}^{t-1} \nabla \mathcal{J}(\theta_k), 
    \label{eq:thetat}
\end{equation}
\vspace{-3mm}
\begin{equation}
    {\theta'}_t = \theta_0 - \alpha \sum_{k=0}^{t-1} (1-\gamma^{t-k-1})\nabla \mathcal{J}(\theta_k).
    \label{eq:thetatp}    
\end{equation}
Comparing (\ref{eq:thetat}) and (\ref{eq:thetatp}), we can see that $\theta'$, the target for the consistency loss in MT, is updated slower than the model parameter $\theta$ because of the use of the EMA decay factor $\gamma$.
On the other hand, in $\Pi$ model, because $\theta' = \theta$, the target is updated faster than that of MT
As described in the supplementary, we can get the same results of slow target update in MT for the SGD with momentum case that we used for our experiments.

Now we will check why MT performs better than $\Pi$ model in CISSL environment.
Assume $\theta^{\Pi}$ and $\theta^{MT}$ be initially with the same value $\theta$. In this case, the consistency loss of $\Pi$ model and MT are 
\begin{equation}
\begin{split}   
    \Pi \ \text{model}: & L_{con}^{\Pi}(\theta) = d(f_\theta(X + \epsilon), f_{\theta'=\theta}(X+\epsilon')) \\
    \text{MT}: & L_{con}^{MT}(\theta) = d(f_{\theta}(X + \epsilon), f_{\theta'}(X+\epsilon')).
    \label{eq:L_cons}
\end{split}
\end{equation}
If we use $L_2$ distance for $d$ for simplicity, their derivatives become  
\begin{equation}
    \begin{split}
        \nabla_\theta L&_{con}^{\Pi} = \nabla_\theta \frac{1}{2} [f_{\theta}(X+\epsilon) - f_{\theta}(X+\epsilon')]^2 \\
        & = [f_{\theta}(X+\epsilon) - f_{\theta}(X+\epsilon')] \nabla_\theta f_{\theta}(X+\epsilon) ,
    \end{split}
    \label{eq:dL_pi}
\end{equation}
\begin{equation}
    \begin{split}
        \nabla_\theta L&_{con}^{MT} =\nabla_\theta \frac{1}{2} [f_{\theta}(X+\epsilon) - f_{\theta'}(X+\epsilon')]^2 \\
        &= [f_{\theta}(X+\epsilon) - f_{\theta'}(X+\epsilon')] \nabla_\theta f_{\theta}(X+\epsilon). 
    \end{split}
    \label{eq:dL_mt}
\end{equation}
Note the target parameters ($\theta'$) in (\ref{eq:L_cons}) are not included in the gradient calculation. 
Using the Taylor series expansion $f_{\theta'}(X+\epsilon') \simeq f_{\theta}(X+\epsilon') + (\theta' - \theta)^T\nabla_\theta f_\theta(X+\epsilon')$ and subtracting (\ref{eq:dL_pi}) from (\ref{eq:dL_mt}), we obtain
\begin{equation}
    \begin{split}
    \nabla&_\theta L_{con}^{MT} - \nabla_\theta L^{\Pi}_{con} \\ &= \nabla_\theta f_\theta(X+\epsilon') (\theta - \theta')^T \nabla_\theta f_\theta(X+\epsilon)\\
    &\simeq \nabla_\theta f_\theta (X) \nabla_\theta f_\theta (X)^T (\theta - \theta').
    \end{split}
    \label{eq:difference}
\end{equation}
In the last line of (\ref{eq:difference}), we assumed gradients be constant in a small area around $X$.
When the sample $X$ is far away from the decision boundary, $\nabla_\theta f_\theta (X) \simeq 0$ and MT and $\Pi$ model behave the same, but in the area near the decision boundary, it becomes $||\nabla_\theta f_\theta (X)|| \gg 0$, and in the gradient descent step, compared to the $\Pi$ model, the negative gradient of MT ($\nabla_\theta \mathcal{J}$ in (\ref{eq:objective_function})) prohibits $\theta$ from being away from the target $\theta'$.
In the CISSL environment, while $\Pi$ model pushes the boundary towards the minor class, MT mitigates this by retaining the old target boundary like ensemble models.

In summary, the performance difference between the $\Pi$ model and MT in CISSL is due to different targets of consistency regularization.
The $\Pi$ model uses the current model ($\theta$) as a target.
Therefore, the model smooths the decision boundary regardless of whether it passes the high-density area of the minor class.
Because the target is the same as the parameter, smoothing causes model degradation as the parameter update is repeated.
MT, on the other hand, targets a more conservative model ($\theta'$) than the current model.
Note that since the target of MT is different from the current model, even if we reduce the learning rate of the $\Pi$ model, it would work differently from MT.
The conservative target has an ensemble effect with consistency regularization, so smoothing does not cause severe performance degradation.

Besides, we can explain the reason why MT performs better than the $\Pi$ model in terms of batch sampling.
In the mini-batch, minor class samples are sampled at a relatively low frequency.
For this reason, the $\Pi$ model frequently updates the model without a minor sample during the consistency regularization, which distorts the decision boundary.
On the other hand, since the target of MT is calculated by EMA, even if there is no minor class sample in the mini-batch, it includes more information about the minor class samples.
Thus, we can say that MT learns with a more stable target than the $\Pi$ model.

\section{Suppressed Consistency Loss}
\label{sec:scl}

In Section \ref{sec:analysis}, we found that the main performance degradation of SSL models in CISSL is due to consistency regularization in minor classes.
With the intuition that we should suppress the consistency regularization of minor classes in CISSL, we propose a new loss term, \textit{suppressed consistency loss} (SCL), as follows:
\begin{equation}
\begin{split}   
    L_{SCL}(X_i) = g(N_c) * L_{con}(X_i),\\
    \text{where} \quad c = \text{argmax}(f_{\theta}(X_i)).
    \label{eq:scl}
\end{split}
\end{equation}
Here, $g(z)$ can be any function inversely proportional to $z$ and we set it as 
\begin{equation}
    g(z) = \beta^{1-\frac{z}{N_{max}}},
    \label{eq:scl_f}
\end{equation}
where $\beta \in (0, 1]$.
$N_c$ is the number of training samples of the class predicted by the model, $N_{max}$ is the number of samples of the class with the most frequency.
SCL weights the consistency loss in an exponentially inverse proportional to the number of samples in a class.
In (\ref{eq:scl}), $g(N_c)$ is 1 for the most frequent class, where it works the same as the conventional consistency loss.
For the least frequent class, the influence of the consistency loss is suppressed.
In (\ref{eq:scl_f}), the exponential decay is to incorporate very high imbalance factor in our model. However, when the imbalance factor is not so high, a simple linear decay can also be used.

Fig.\ref{fig:algorithm} illustrates the effect of consistency regularization by SCL.
When training with SCL, the decision boundary is smoothed weakly for minor class and is smoothed strongly for major class.
If the performance of the model is inaccurate, especially for the minor class, it may pass through the high-density area.
Then the SCL limits the smoothing of the decision boundary towards the minor class cluster.
On the other hand, when the model mispredicts actual minor class samples as a major class in the high-density area of the minor class, the decision boundary is smoothed with higher weight.
Consequentially, SCL pushes the decision boundary to low-density areas of the minor class and prevents performance degradation, as shown in Fig.\ref{fig:algorithm}.

\begin{table*}[ht]
    \centering
    \renewcommand{\tabcolsep}{0.7mm}
    \caption{Test error rates~(\%) from experiments with 4k number of labeled data and imbalance factor \{10, 20, 50, 100 \} under 3 different unlabeled imbalance types in CIFAR10 and imbalance factor \{10, 20, 50, 100 \} under 3 different unlabeled imbalance types in SVHN.
    VAT+EM refers to Virtual Adversarial Training with Entropy Minimization.
    To improve legibility, the standard deviation is listed in supplemental materials.
    (\textbf{Bold}/\textcolor{red}{\textbf{Red}}/\textcolor{blue}{\textbf{Blue}}: \textbf{supervised}, \textcolor{red}{\textbf{best}} and \textcolor{blue}{\textbf{second best}} results for each column.)
    }
    \label{tab:imbalance_factor}
    \begin{center}
    \begin{small}
    \begin{sc}
    \adjustbox{max width=\linewidth}{
    \begin{subtable}{\linewidth}
    \centering
    \vspace{-4mm}
    \caption{CIFAR10}
    \vspace{-2mm}
    \label{tab:cifar_imbalance_factor}
    \begin{tabularx}{\linewidth}{l||r|r|r|r||r|r|r|r||r|r|r|r}
        \toprule
        Unlabel Imbalance Type & \multicolumn{4}{c||}{Uniform ($\rho_u=1$)} & \multicolumn{4}{c||}{Half ($\rho_u = \rho_l/2$)} & \multicolumn{4}{c}{Same ($\rho_u=\rho_l$)} \\
        \midrule
        Imbalance factor ($\rho_l$) & \multicolumn{1}{c|}{10} & \multicolumn{1}{c|}{20} & \multicolumn{1}{c|}{50} & \multicolumn{1}{c||}{100} & \multicolumn{1}{c|}{10} & \multicolumn{1}{c|}{20} & \multicolumn{1}{c|}{50} & \multicolumn{1}{c||}{100} & \multicolumn{1}{|c|}{10} & \multicolumn{1}{c|}{20} & \multicolumn{1}{c|}{50} & \multicolumn{1}{c}{100}  \\
        \midrule
Supervised & \textbf{23.03} & \textbf{27.49} & \textbf{33.15} & \textbf{36.71} & \textbf{23.03} & \textbf{27.49} & \textbf{33.15} & \textbf{36.71} & \textbf{23.03} & \textbf{27.49} & \textbf{33.15} & \textbf{36.71} \\
$\Pi$-Model \cite{laine2016temporal} & 21.10 & 25.74 & 33.91 & 39.36 & 22.69 & 27.72 & 33.96 & 38.84 & 23.49 & 28.18 & 34.22 & 38.05 \\
MT \cite{tarvainen2017mean} & \textcolor{blue}{\textbf{16.45}} & \textcolor{blue}{\textbf{19.25}} & \textcolor{blue}{\textbf{23.45}} & \textcolor{blue}{\textbf{29.06}} & \textcolor{blue}{\textbf{19.48}} & \textcolor{blue}{\textbf{23.30}} & \textcolor{blue}{\textbf{30.06}} & \textcolor{blue}{\textbf{35.37}} & 20.50 & \textcolor{blue}{\textbf{24.67}} & \textcolor{blue}{\textbf{31.77}} & \textcolor{blue}{\textbf{35.91}} \\
VAT + em \cite{miyato2018virtual} & 17.93 & 20.18 & 30.43 & 36.57 & 20.17 & 24.50 & 32.54 & 36.77 & 21.45 & 25.83 & 33.13 & 37.67 \\
VAT + em + SNTG \cite{luo2018smooth} & 18.15 & 20.39 & 29.77 & 36.34 & 20.41 & 24.64 & 32.56 & 38.48 & 21.87 & 26.49 & 33.36 & 38.48 \\
Pseudo-Label \cite{lee2013pseudo} & 19.33 & 24.34 & 34.18 & 39.59 & 21.23 & 26.78 & 34.12 & 39.72 & 22.73 & 27.50 & 34.91 & 38.69 \\
ICT \cite{verma2019interpolation} & 18.01 & 20.52 & 30.18 & 38.33 & 19.53 & 23.90 & 31.09 & 37.36 & \textcolor{blue}{\textbf{19.96}} & 25.63 & 33.56 & 36.85 \\
MT+SCL (ours) & \textcolor{red}{\textbf{15.65}} & \textcolor{red}{\textbf{16.99}} & \textcolor{red}{\textbf{19.95}} & \textcolor{red}{\textbf{22.62}} & \textcolor{red}{\textbf{17.36}} & \textcolor{red}{\textbf{21.74}} & \textcolor{red}{\textbf{28.20}} & \textcolor{red}{\textbf{33.09}} & \textcolor{red}{\textbf{18.69}} & \textcolor{red}{\textbf{22.98}} & \textcolor{red}{\textbf{29.76}} & \textcolor{red}{\textbf{34.22}} \\
         \bottomrule
    \end{tabularx}
    \end{subtable}%
    }
    \adjustbox{max width=\linewidth}{
    \begin{subtable}{\linewidth}
    \centering
    \bigskip
    \vspace{-2mm}
    \caption{SVHN}
    \vspace{-2mm}
    \label{tab:svhn_imbalance_factor}
    \begin{tabularx}{\linewidth}{l||r|r|r|r||r|r|r|r||r|r|r|r}
        \toprule
        Unlabel Imbalance Type & \multicolumn{4}{c||}{Uniform ($\rho_u=1$)} & \multicolumn{4}{c||}{Half ($\rho_u = \rho_l/2$)} & \multicolumn{4}{c}{Same ($\rho_u=\rho_l$)} \\
        \midrule
        Imbalance factor ($\rho_l$) & \multicolumn{1}{c|}{10} & \multicolumn{1}{c|}{20} & \multicolumn{1}{c|}{50} & \multicolumn{1}{c||}{100} & \multicolumn{1}{c|}{10} & \multicolumn{1}{c|}{20} & \multicolumn{1}{c|}{50} & \multicolumn{1}{c||}{100} & \multicolumn{1}{|c|}{10} & \multicolumn{1}{c|}{20} & \multicolumn{1}{c|}{50} & \multicolumn{1}{c}{100}  \\
        \midrule
Supervised & \textbf{18.49}  & \textbf{21.92}  & \textbf{30.03}  & \textbf{35.89} & \textbf{18.49}  & \textbf{21.92}  & \textbf{30.03}  & \textbf{35.89}  & \textbf{18.49}  & \textbf{21.92}  & \textbf{30.03}  & \textbf{35.89}  \\
$\Pi$-Model \cite{laine2016temporal} & 11.74  & 13.42  & 21.63  & 28.59  & 12.96  & 16.70  & 24.02  & 33.73  & 13.46  & 17.13  & 26.53  & 33.71  \\
MT \cite{tarvainen2017mean} & \textcolor{red}{\textbf{6.52}}  & \textcolor{red}{\textbf{6.75}}  & \textcolor{red}{\textbf{7.60}}  & \textcolor{blue}{\textbf{8.94}}  & \textcolor{red}{\textbf{7.25}}  & \textcolor{red}{\textbf{8.85}}  & \textcolor{blue}{\textbf{12.19}}  & \textcolor{red}{\textbf{17.23}}  & \textcolor{blue}{\textbf{8.62}}  & \textcolor{red}{\textbf{9.29}}  & \textcolor{red}{\textbf{15.16}}  & \textcolor{blue}{\textbf{21.01}}  \\
VAT + em \cite{miyato2018virtual} & 6.81  & 7.70  & 13.84  & 29.15  & 8.99  & 11.59  & 18.95  & 30.44  & 10.39  & 13.62  & 21.49  & 32.39  \\
VAT + em + SNTG \cite{luo2018smooth} & 93.30  & 93.30  & 14.88  & 93.30  & 93.30  & 93.30  & 20.60  & 93.30  & 93.30  & 93.30  & 23.52  & 93.30  \\
Pseudo-Label \cite{lee2013pseudo} & 10.15  & 9.97  & 16.00  & 32.79  & 11.59  & 13.97  & 24.40  & 33.70  & 12.34  & 15.93  & 25.66  & 33.53  \\
ICT \cite{verma2019interpolation} & 27.82  & 37.75  & 58.20  & 67.02  & 22.38  & 38.12  & 48.88  & 58.99  & 24.53  & 37.25  & 49.85  & 56.97  \\
MT+SCL (ours) & \textcolor{red}{\textbf{6.52}}  & \textcolor{blue}{\textbf{7.11}}  & \textcolor{blue}{\textbf{7.70}}  & \textcolor{red}{\textbf{8.56}}  & \textcolor{blue}{\textbf{7.54}}  & \textcolor{blue}{\textbf{9.29}}  & \textcolor{red}{\textbf{11.46}}  & \textcolor{blue}{\textbf{18.63}}  & \textcolor{red}{\textbf{8.22}}  & \textcolor{blue}{\textbf{10.04}}  & \textcolor{blue}{\textbf{15.48}}  & \textcolor{red}{\textbf{20.39}}  \\

         \bottomrule
    \end{tabularx}
    \end{subtable}
    }
    \end{sc}
    \end{small}
    \end{center}
    \vspace{-6mm}
\end{table*}

\section{Experiments}
\label{sec:experiments}

\subsection{Datasets and implementation details}
\label{exp:dataset}

We conducted experiments using the CIFAR10 \cite{krizhevsky2009learning} and SVHN \cite{netzer2011reading} datasets in our proposed environments and followed the common practice in SSL and CIL \cite{oliver2018realistic, johnson2019survey}. 
We divided the training dataset into three parts: labeled dataset, unlabeled dataset, and validation dataset.
Labeled data is configured to have an imbalance for each class according to the CIL environment.
We have experimented with various numbers of labeled samples and imbalance factors.
We considered three types of class imbalance in unlabeled data:
\textit{Same} ($\rho_u = \rho_l$, where $\rho_l$ and $\rho_u$ are the imbalance factors for labeled and unlabeled dataset.), \textit{Uniform} (uniform distribution, $\rho_u=1$), and \textit{Half} ($\rho_u = \rho_l/2$).
The size of the unlabeled dataset changes depending on unlabeled data imbalance types because of the limitation of the dataset used.
For fair experiments, we set the size of the unlabeled set based on the \textit{Same} case, which uses the lowest number of unlabeled samples.
Fig.\ref{fig:type} shows the three imbalance types with imbalance factor 10.
Validation data is made up as in \cite{oliver2018realistic}.

In all experiments, we used the Wide-Resnet-28-2 model~\cite{zagoruyko2016wide}.
It has enough capacity to show the performance improvement of SSL objectively~\cite{oliver2018realistic}, and it is used in the new SSL methods \cite{berthelot2019mixmatch, verma2019interpolation}.
We adopt optimizer and learning rate from \cite{verma2019interpolation}, and other hyper-parameters are set under a similar setting with \cite{oliver2018realistic}\footnote{https://github.com/brain-research/realistic-ssl-evaluation}.
In our experiments, we used third-party implementation \footnote{https://github.com/perrying/realistic-ssl-evaluation-pytorch}.
All the scores of test error rates are from five independent runs with different random seeds.
Experiments with different random seeds shuffle the frequency ranking of each class when the imbalance factor is constant, and cover a variety of cases. 

\begin{table*}[ht]
    \centering
    \renewcommand{\tabcolsep}{0.7mm}
    \caption{Test error rates~(\%) from experiments with different re-weighting methods in CIFAR10 and SVHN. We compared inverse and normalization (IN), focal loss (FOCAL), and class-balanced loss (CB) to conventional cross-entropy loss (CE).\\
    (\textcolor{red}{\textbf{Red}}: best results for each row with same unlabeled data imbalance.)
    }
    \vspace{-4mm}
    \label{tab:reweight}
    \begin{center}
    \begin{small}
    \begin{sc}
    \adjustbox{max width=\linewidth}{
    \begin{subtable}{\linewidth}
    \centering
    \caption{CIFAR10}
    \vspace{-2mm}
    \label{tab:cifar_reweight}
    \begin{tabularx}{1.03\linewidth}{l||r|r|r|r||r|r|r|r||r|r|r|r}
        \toprule
        Unlabel Imbalance Type & \multicolumn{4}{c||}{Uniform ($\rho_u=1$)} & \multicolumn{4}{c||}{Half ($\rho_u = \rho_l/2$)} & \multicolumn{4}{c}{Same ($\rho_u=\rho_l$)} \\
         Re-weighting Method & \multicolumn{1}{c|}{CE} & \multicolumn{1}{c|}{IN} & \multicolumn{1}{c|}{Focal} & \multicolumn{1}{c||}{CB} & \multicolumn{1}{c|}{CE} & \multicolumn{1}{c|}{IN} & \multicolumn{1}{c|}{Focal} & \multicolumn{1}{c||}{CB} & \multicolumn{1}{c|}{CE} & \multicolumn{1}{c|}{IN} & \multicolumn{1}{c|}{Focal} & \multicolumn{1}{c}{CB} \\
         \midrule
Supervised & 36.71  & \textcolor{red}{\textbf{35.73}} & 36.80  & 37.19  & 36.71  & \textcolor{red}{\textbf{35.73}} & 36.80  & 37.19  & 36.71  & \textcolor{red}{\textbf{35.73}} & 36.80  & 37.19  \\
$\Pi$-Model \cite{laine2016temporal} & 39.36  & \textcolor{red}{\textbf{36.90}} & 39.89  & 39.20  & 38.84  & 37.82  & 38.28  & \textcolor{red}{\textbf{37.51}} & 38.05  & \textcolor{red}{\textbf{37.18}} & 38.10  & 37.34  \\
MT \cite{tarvainen2017mean} & 29.06  & \textcolor{red}{\textbf{24.00}} & 30.73  & 29.50  & 35.37  & \textcolor{red}{\textbf{33.08}} & 34.45  & 35.04  & 35.91  & \textcolor{red}{\textbf{34.01}} & 35.65  & 35.17  \\
VAT + em \cite{miyato2018virtual} & 36.57  & \textcolor{red}{\textbf{31.34}} & 37.51  & 36.78  & 36.77  & \textcolor{red}{\textbf{36.20}} & 37.62  & 38.13  & 37.67  & \textcolor{red}{\textbf{36.91}} & 37.88  & 37.64  \\
VAT + em + SNTG \cite{luo2018smooth} & 36.34  & \textcolor{red}{\textbf{33.03}} & 37.78  & 36.26  & 38.48  & \textcolor{red}{\textbf{35.90}} & 38.01  & 37.44  & 38.48  & \textcolor{red}{\textbf{36.99}} & 37.71  & 37.53  \\
Pseudo-Label \cite{lee2013pseudo} & 39.59  & \textcolor{red}{\textbf{30.62}} & 37.90  & 39.38  & 39.72  & \textcolor{red}{\textbf{37.36}} & 38.77  & 39.12  & 38.69  & \textcolor{red}{\textbf{36.84}} & 38.92  & 38.52  \\
MT+SCL (ours) & 22.62  & \textcolor{red}{\textbf{21.59}} & 23.44  & 22.93  & 33.09  & \textcolor{red}{\textbf{31.63}} & 34.09  & 33.14  & 34.22  & \textcolor{red}{\textbf{32.09}} & 33.93  & 34.66  \\
         \bottomrule
    \end{tabularx}
    \end{subtable}%
    }
    \adjustbox{max width=\linewidth}{
    \begin{subtable}{\linewidth}
    \centering
    \bigskip
    \vspace{-2mm}
    \caption{SVHN}
    \vspace{-2mm}
    \label{tab:svhn_reweight}
    \begin{tabularx}{1.03\linewidth}{l||r|r|r|r||r|r|r|r||r|r|r|r}
        \toprule
        Unlabel Imbalance Type & \multicolumn{4}{c||}{Uniform ($\rho_u=1$)} & \multicolumn{4}{c||}{Half ($\rho_u = \rho_l/2$)} & \multicolumn{4}{c}{Same ($\rho_u=\rho_l$)} \\
         Re-weighting Method & \multicolumn{1}{c|}{CE} & \multicolumn{1}{c|}{IN} & \multicolumn{1}{c|}{Focal} & \multicolumn{1}{c||}{CB} & \multicolumn{1}{c|}{CE} & \multicolumn{1}{c|}{IN} & \multicolumn{1}{c|}{Focal} & \multicolumn{1}{c||}{CB} & \multicolumn{1}{c|}{CE} & \multicolumn{1}{c|}{IN} & \multicolumn{1}{c|}{Focal} & \multicolumn{1}{c}{CB} \\
         \midrule
Supervised & 35.89  & \textcolor{red}{\textbf{34.60}} & 35.45  & 35.30  & 35.89  & \textcolor{red}{\textbf{34.60}} & 35.45  & 35.30  & 35.89  & \textcolor{red}{\textbf{34.60}} & 35.45  & 35.30  \\
$\Pi$-Model \cite{laine2016temporal} & 28.59  & \textcolor{red}{\textbf{26.72}} & 30.15  & 27.99  & 33.73  & \textcolor{red}{\textbf{29.60}} & 31.67  & 31.12  & 33.71  & \textcolor{red}{\textbf{31.70}} & \textcolor{red}{\textbf{31.70}} & 33.17  \\
MT \cite{tarvainen2017mean} & 8.94  & \textcolor{red}{\textbf{6.82}} & 8.66  & 7.86  & 17.23  & 17.02  & \textcolor{red}{\textbf{16.20}} & 16.68  & 21.01  & 20.80  & \textcolor{red}{\textbf{20.01}} & 21.77  \\
VAT + em \cite{miyato2018virtual} & 29.15  & \textcolor{red}{\textbf{20.26}} & 28.09  & 29.37  & 30.44  & \textcolor{red}{\textbf{27.44}} & 28.62  & 29.65  & 32.39  & \textcolor{red}{\textbf{29.18}} & 30.62  & 30.93  \\
VAT + em + SNTG \cite{luo2018smooth} & 93.30  & 93.30  & 93.30  & 93.30  & 93.30  & 93.30  & 93.30  & 93.30  & 93.30  & 93.30  & 93.30  & 93.30  \\
Pseudo-Label \cite{lee2013pseudo} & 32.79  & \textcolor{red}{\textbf{13.48}} & 35.07  & 34.38  & 33.70  & \textcolor{red}{\textbf{31.83}} & 32.79  & 32.83  & 33.53  & \textcolor{red}{\textbf{31.62}} & 33.63  & 34.55  \\
MT+SCL (ours) & 8.56  & 8.48  & \textcolor{red}{\textbf{7.74}} & 9.02  & 18.63  & 18.59  & \textcolor{red}{\textbf{16.34}} & 16.44  & \textcolor{red}{\textbf{20.39}} & 20.51  & 20.95  & 21.06  \\
         \bottomrule
    \end{tabularx}
    \end{subtable}
    }
    \end{sc}
    \end{small}
    \end{center}
    \vspace{-4mm}
\end{table*}

\begin{table*}[ht]
    \centering
    \renewcommand{\tabcolsep}{0.7mm}
    \caption{Test error rates~(\%) from experiments with imbalance factor 100 and the number of labeled data \{1k, 2k, 4k\} under 3 different unlabeled imbalance types in CIFAR10 and the number of labeled data \{250, 500, 1k\} under 3 different unlabeled imbalance types in SVHN. 
    Details are the same as Table \ref{tab:imbalance_factor}.
    }
    \vspace{-5mm}
    \label{tab:num_of_label}
    \begin{center}
    \begin{small}
    \begin{sc}
    \adjustbox{max width=\linewidth}{
    \begin{subtable}{\linewidth}
    \centering
    \caption{CIFAR10}
    \vspace{-2mm}
    \label{tab:cifar_num_of_label}
    \begin{tabularx}{.83\linewidth}{l||r|r|r||r|r|r||r|r|r}
        \toprule
        Unlabel Imbalance Type & \multicolumn{3}{c||}{Uniform ($\rho_u=1$)} & \multicolumn{3}{c||}{Half ($\rho_u = \rho_l/2$)} & \multicolumn{3}{c}{Same ($\rho_u=\rho_l$)} \\
        \midrule
         \# labeled data & \multicolumn{1}{c|}{1000} & \multicolumn{1}{c|}{2000} & \multicolumn{1}{c||}{4000} & \multicolumn{1}{c|}{1000} & \multicolumn{1}{c|}{2000} & \multicolumn{1}{c||}{4000} & \multicolumn{1}{c|}{1000} & \multicolumn{1}{c|}{2000} & \multicolumn{1}{c}{4000}\\
         \midrule
Supervised & \textbf{54.24}  & \textbf{45.81}  & \textbf{36.71}  & \textbf{54.24}  & \textbf{45.81}  & \textbf{36.71}  & \textbf{54.24}  & \textbf{45.81}  & \textbf{36.71}  \\
$\Pi$-Model \cite{laine2016temporal} & 56.82  & 48.55  & 39.36  & 55.99  & 47.74  & 38.84  & 55.42  & 46.83  & 38.05  \\
MT \cite{tarvainen2017mean} & \textcolor{blue}{\textbf{51.74}}  & \textcolor{blue}{\textbf{38.94}}  & \textcolor{blue}{\textbf{29.06}}  & \textcolor{blue}{\textbf{51.61}}  & \textcolor{blue}{\textbf{42.47}}  & \textcolor{blue}{\textbf{35.37}}  & \textcolor{blue}{\textbf{52.58}}  & \textcolor{blue}{\textbf{44.11}}  & \textcolor{blue}{\textbf{35.91}}  \\
VAT + em \cite{miyato2018virtual} & 53.68  & 48.47  & 36.57  & 53.60  & 45.20  & 36.77  & 53.62  & 44.77  & 37.67  \\
VAT + em + SNTG \cite{luo2018smooth} & 54.53  & 48.23  & 36.34  & 55.59  & 45.37  & 38.48  & 55.55  & 45.99  & 38.48  \\
Pseudo-Label \cite{lee2013pseudo} & 58.19  & 50.01  & 39.59  & 57.05  & 49.42  & 39.72  & 56.68  & 48.45  & 38.69  \\
ICT \cite{verma2019interpolation} & 57.10  & 48.25  & 38.33  & 56.02  & 47.60  & 37.36  & 55.10  & 47.19  & 36.85  \\
MT+SCL (ours) & \textcolor{red}{\textbf{42.84}}  & \textcolor{red}{\textbf{28.69}}  & \textcolor{red}{\textbf{22.62}}  & \textcolor{red}{\textbf{45.72}}  & \textcolor{red}{\textbf{39.97}}  & \textcolor{red}{\textbf{33.09}}  & \textcolor{red}{\textbf{48.00}}  & \textcolor{red}{\textbf{40.69}}  & \textcolor{red}{\textbf{34.22}}  \\
         \bottomrule
    \end{tabularx}
    \end{subtable}%
    }
    \adjustbox{max width=\linewidth}{
    \begin{subtable}{\linewidth}
    \centering
    \bigskip
    \vspace{-2mm}
    \caption{SVHN}
    \vspace{-2mm}
    \label{tab:svhn_num_of_label}
        \begin{tabularx}{.83\linewidth}{l||r|r|r||r|r|r||r|r|r}
        \toprule
        Unlabel Imbalance Type & \multicolumn{3}{c||}{Uniform ($\rho_u=1$)} & \multicolumn{3}{c||}{Half ($\rho_u = \rho_l/2$)} & \multicolumn{3}{c}{Same ($\rho_u=\rho_l$)} \\
        \midrule
         \# labeled data & \multicolumn{1}{c|}{250} & \multicolumn{1}{c|}{500} & \multicolumn{1}{c||}{1000} & \multicolumn{1}{c|}{250} & \multicolumn{1}{c|}{500} & \multicolumn{1}{c||}{1000} & \multicolumn{1}{c|}{250} & \multicolumn{1}{c|}{500} & \multicolumn{1}{c}{1000}\\
         \midrule
Supervised & \textbf{61.31}  & \textbf{47.98}  & \textbf{35.89}  & \textbf{61.31}  & \textbf{47.98}  & \textbf{35.89}  & \textbf{61.31}  & \textbf{47.98}  & \textbf{35.89}  \\
$\Pi$-Model \cite{laine2016temporal} & 54.51  & 39.49  & 28.59  & 54.14  & 42.20  & 33.73  & 54.10  & 43.89  & 33.71  \\
MT \cite{tarvainen2017mean} & \textcolor{blue}{\textbf{38.32}}  & \textcolor{blue}{\textbf{18.14}}  & \textcolor{blue}{\textbf{8.94}}  & \textcolor{blue}{\textbf{41.72}}  & \textcolor{blue}{\textbf{23.33}}  & \textcolor{red}{\textbf{17.23}}  & \textcolor{blue}{\textbf{42.42}}  & \textcolor{blue}{\textbf{28.86}}  & \textcolor{blue}{\textbf{21.01}}  \\
VAT + em \cite{miyato2018virtual} & 64.67  & 44.04  & 29.15  & 58.01  & 41.15  & 30.44  & 55.03  & 42.44  & 32.39  \\
VAT + em + SNTG \cite{luo2018smooth} & 65.02  & 93.30  & 93.30  & 57.94  & 93.30  & 93.30  & 54.19  & 93.30  & 93.30  \\
Pseudo-Label \cite{lee2013pseudo} & 63.16  & 49.78  & 32.79  & 54.79  & 44.32  & 33.70  & 56.83  & 43.71  & 33.53  \\
ICT \cite{verma2019interpolation} & 86.54  & 77.64  & 67.02  & 84.22  & 72.21  & 58.99  & 85.15  & 71.19  & 56.97  \\
MT+SCL (ours) & \textcolor{red}{\textbf{26.25}}  & \textcolor{red}{\textbf{15.31}}  & \textcolor{red}{\textbf{8.56}}  & \textcolor{red}{\textbf{33.44}}  & \textcolor{red}{\textbf{22.26}}  & \textcolor{blue}{\textbf{18.63}}  & \textcolor{red}{\textbf{35.32}}  & \textcolor{red}{\textbf{27.13}}  & \textcolor{red}{\textbf{20.39}}  \\
         \bottomrule
    \end{tabularx}
    \end{subtable}
    }
    \end{sc}
    \end{small}
    \end{center}
\vspace{-5mm}
\end{table*}
 
\subsection{Baselines to CISSL}
\label{exp:baseline}

We conducted experiments on how existing methods in the field of SSL and CIL perform in our defined CISSL environment and used them as the baseline for our research.
We experimented in the case of 4k and 1k labeled samples for CIFAR10 and SVHN each, both with imbalance factor 100.

\noindent \textbf{A. Comparison of Semi-supervised Learning Methods} \\
Columns with imbalance factor 100 in Table.\ref{tab:cifar_imbalance_factor} is the results of applying the SSL methods to the CISSL problem in CIFAR10.
Except for MT, almost all SSL methods are inferior to supervised learning.
Even if the unlabeled data imbalance is mitigated to \textit{Uniform} case, there is no improvement in the performance of SSL methods except MT.

Columns with imbalance factor 100 in Table.\ref{tab:svhn_imbalance_factor} is the same experiment for SVHN.
Most SSL methods perform better when the unlabeled data imbalance is lower, i.e. in \textit{Uniform} case than in \textit{Same} case.
Notably, ICT showed a performance degradation of over $21\%p$ compared to the supervised learning, and SNTG even failed to train a model.

From this experimental results and the analysis in Section.\ref{sec:analysis}, we used MT as our baseline, which performed best in all experiments.

\noindent \textbf{B. Comparison of Class Imbalanced Learning Methods} \\
We carried out the ablation experiments to cross-entropy loss (CE) as three types of CIL: Inverse and Normalization (IN), Focal loss, and Class-Balanced (CB) loss.
We applied these CIL methods only to the supervised loss, $L_{sup}$ in (\ref{eq:objective_function}), and did not apply them to unlabeled data because we do not know the class label of the unlabeled data.
In this experiment, we ignored ICT because CIL methods cannot be applied to ICT which uses mixup supervised loss.

Table.\ref{tab:cifar_reweight} is the result of CIFAR10 experimented with imbalance factor 100, 4k labeled dataset.
First of all, it seems that not all CIL methods always improve performance over CE.
As unlabeled data imbalance and SSL methods change, their relative performance with CE differs.
In this table, IN shows the best performance in all cases except the \textit{Half} case of the $\Pi$ model.

Table.\ref{tab:svhn_reweight} is the result of SVHN experiments with imbalance factor 100, 1k labeled dataset.
Unlike the previous CIFAR10 results, IN does not always dominate.
The best algorithm differs according to the unlabeled data imbalance type in MT and our method.
Since we do not know the unlabeled data imbalance beforehand, choosing a specific CIL algorithm does not guarantee a performance boost.
So we used the most common cross-entropy as our baseline.
In addition, SNTG failed to learn, as in Table.\ref{tab:svhn_imbalance_factor}.

\subsection{Unlabeled data Imbalance}
\label{exp:unlabel_imb}

\noindent \textbf{A. Comparison of Imbalance Factor}\\
We experimented with changing the imbalance factor while keeping the number of labeled samples.
We experimented on CIFAR-10 and SVHN with imbalance factor $\rho_l \in \{10, 20, 50, 100\}$.
The results are shown in Table \ref{tab:cifar_imbalance_factor}, \ref{tab:svhn_imbalance_factor}, respectively.

In Table \ref{tab:cifar_imbalance_factor}, the higher the imbalance factor, the lower the performance.
Supervised learning on imbalance factor 100 achieves $36.71\%$ error, which $13\%p$ higher than supervised learning on imbalance factor 10.
In the case of the small imbalance factor, SSL algorithms generally improve performance although unlabeled data has same imbalance with labeled data.
As the imbalance factor increases, on the other hand, some SSL algorithms show lower performance than supervised learning.
Mean Teacher is the only SSL algorithm that improves the performance with imbalance factor 100 in \textit{Same} case.
This means that general SSL algorithms do not consider the imbalance for the unlabeled data.
However, the proposed SCL has robustly improved the performance in various imbalance settings.
Notably, it shows remarkable improvement in the \textit{Uniform} case compared to SSL algorithms.

Table.\ref{tab:svhn_imbalance_factor} shows similar results.
However, there is no big performance difference between MT and our method.
This is because SVHN is easier to classify than CIFAR10.
For SVHN, SNTG and ICT show lower performance than the supervised learning.
It seems that the model training fails.
We discuss this phenomenon in Section.\ref{sec:discussion}.

\noindent \textbf{B. Comparison of The Number of Labeled Samples}\\
We experimented with keeping the imbalance factor while changing the number of labeled samples.
We set the number of labeled data to \{1k, 2k, 4k\} in CIFAR10, and \{250, 500, 1k\} in SVHN.
The results of CIFAR10 and SVHN are shown in Table \ref{tab:cifar_num_of_label}, \ref{tab:svhn_num_of_label}, respectively.

In Table.\ref{tab:cifar_num_of_label}, the smaller the size of the labeled set, the lower the performance.
In particular, when the size of the labeled data is 1k, most of the algorithms are weaker than supervised learning , while our method improves performance.
This result indicates that consistency regularization is not valid when the baseline classifier is not performing well.

Table.\ref{tab:svhn_num_of_label} also shows similar tendency between the size of labeled data and performance. For SNTG and ICT, same as Section.\ref{exp:unlabel_imb}.A, they have lower performance than supervised learning, either.

\begin{table}[ht]
    \centering
    \vspace{-3mm}
    \renewcommand{\tabcolsep}{0.7mm}
    \caption{Detection results for PASCAL VOC2007 testset. 
    cls and loc are the consistency loss for classification and localization, respectively. 
    We trained SSD300 on VOC07(L)+VOC12(U). Our result is from three independent trials.
    }
    \label{tab:object_detection}
    \begin{center}
    \begin{small}
    \adjustbox{max width=\linewidth}{
    \begin{tabularx}{1.25\linewidth}{c||c||c|c||c|c}
        \toprule
        Algorithm & Supervised & \multicolumn{2}{c||}{CSD~\cite{jeong2019consistency}} & \multicolumn{2}{c}{CSD + SCL(Ours)} \\
        \hline
        cls &  & o & o & o & o \\
        loc &  &   & o &   & o \\ 
        \hline
         mAP $(\%)$ & 70.2 & 71.7 & 72.3 & 72.07 $\pm$ 0.15 & \textcolor{red}{\bf{72.60 $\pm$ 0.10}} \\
         \bottomrule
    \end{tabularx}
    }
    \end{small}
    \end{center}
    \vspace{-5mm}
\end{table}

\subsection{Object detection}
\label{exp:od}

We followed the CSD \cite{jeong2019consistency} experiment settings and used the SSD300 model \cite{liu2016ssd}.
We used PASCAL VOC2007 trainval dataset as the labeled data and PASCAL VOC2012 trainval dataset.
We evaluated with PASCAL VOC2007 test dataset.
In this experiment, the imbalance factor of labeled dataset is about 20.
We applied our algorithm only to the classification consistency loss of CSD.
The details are in the supplementary material.

In Table~\ref{tab:object_detection}, supervised learning using VOC2007 shows 70.2 mAP.
CSD with only classification consistency loss is 1.5\%p higher than the supervised and CSD shows 2.1\%p of enhancement.
When SCL is applied to the CSD, our method shows additional improvement.

\section{Discussion}
\label{sec:discussion}

The reason why the existing SSL methods did not perform well in the CISSL environment was that they did not consider data imbalance.
This fact gives us some implications.
First, for deep learning to become a practical application, we need to work on a harsher benchmark.
We experimented on datasets which relaxed the equal class distribution assumption of SSL, and our method yielded meaningful results.
Second, we should avoid developing domain-specific algorithms which work very well only under certain conditions.
SNTG~\cite{luo2018smooth} and ICT~\cite{verma2019interpolation} are very good algorithms for existing SSL settings.
In our experiments, however, both algorithms were not robust against class imbalance.
Finally, we need to focus not only on the performance improvement of the model but also on its causes.
An in-depth analysis of the causes of the phenomena provides an intuition about the direction of future research.
Concerning this, we discussed aspects of learning in the CISSL environment in Section.\ref{sec:analysis}.

\section{Conclusion}
\label{sec:conclusion}

In this paper, we proposed Class-Imbalanced Semi-Supervised Learning, which is one step beyond the limitations of SSL.
We theoretically analyzed how the existing SSL methods work in CISSL.
Based on the intuition obtained here, we proposed Suppressed Consistency Loss that works robustly in CISSL.
Our experiments show that our method works well in the CISSL environment compared to the existing SSL and CIL methods, as well as the feasibility of working in object detection.
However, our research have focused on relatively small datasets.
Applying CISSL to more massive datasets would be the future work.

\newpage
\bibliography{main}
\bibliographystyle{icml2020}

\newpage
\twocolumn[
\icmltitle{Supplementary Materials}
]

\setcounter{section}{0}
\renewcommand*{\thesection}{\Alph{section}}

\section{Toy Examples Details}

We generated \textit{two moons} and \textit{four spins} datasets.
We split the train set into labeled data and unlabeled data with imbalance factor 5.
The class distribution of unlabeled data follows \textit{same} case.
The size of the labeled data is 12 (\{2, 10\} samples each) in two moons, 11 (\{1,2,3,5\} samples each) in four spins.
The size of the unlabeled data is 3000 in two moons, 2658 in four spins.
Both datasets have 6,000 validation samples.
We trained each algorithm by 5,000 iterations.
The model is a 3-layer network; optimizer is SGD with momentum, the learning rate is 0.1 decaying at 4,000 iterations multiplied by 0.2, and momentum is 0.9.

In the experiment, we set the function $g(z)$ of suppressed consistency loss as $z / N_{max}$ with simplicity, where $N_c$ is the number of training samples of the class predicted by the model, $N_{max}$ is the number of samples of the class with the most frequency.


\section{$\Pi$ model vs. Mean Teacher Details}
\subsection{SGD Case}
\label{sup:sgd}

Consider the model parameter $\theta$, the learning rate $\alpha$, and the objective function $\mathcal{J}$, then the update rule of SGD optimizer is:
\begin{gather}
    \theta \leftarrow \theta - \alpha \nabla \mathcal{J}(\theta)
\end{gather}
For a EMA decay factor of MT, $\gamma$, the current ($\theta$) and the target ($\theta'$) model parameters at the $t$-th iteration are
\begin{equation}
    \begin{split}
    \theta_t & = \theta_{t-1} - \alpha \nabla \mathcal{J}(\theta_{t-1}) = \dots \\
    & =  \theta_0 - \alpha \sum_{k=0}^{t-1} \nabla \mathcal{J}(\theta_k) \\
    \\
    {\theta'}_t & = \gamma {\theta'}_{t-1} + (1-\gamma) \theta_t  = \dots \\
    & =\theta_0 - \alpha \sum_{k=0}^{t-1} (1-\gamma^{t-k-1})\nabla \mathcal{J}(\theta_k)
    \end{split}
\end{equation}

\subsection{SGD with Momentum Case}
\label{sup:sgdm}

$v$ is the momentum for SGD optimizer, $\delta \in (0,1]$ is a decay factor of momentum, and other parameters are the same with Section.\ref{sup:sgd}.

\begin{gather}
    v \leftarrow \delta v + \alpha \nabla \mathcal{J}(\theta_{t-1}) \\
    \theta \leftarrow \theta - v
\end{gather}

The current model parameter($\theta$) at the $t$-th iteration is

\begin{equation}
    \begin{split}
    \theta_t & = \theta_{t-1} - \delta v_{t-1} + \alpha \nabla \mathcal{J}(\theta_{t-1}) = \dots \\
    & = \theta_0 - \alpha \sum_{k=0}^{t-1} \frac{1-\delta^{t-k}}{1-\delta} \nabla \mathcal{J}(\theta_k)
    \end{split}
\end{equation}

And then the target model($\theta'$) at the $t$-th iteration is
\begin{equation}
    \begin{split}
    {\theta'}_t & = \gamma {\theta'}_{t-1} + (1-\gamma) \theta_t = \dots \\
    & = \theta_0 - \alpha (1-\gamma) \sum_{j=1}^{t} \gamma^{t-j} \{ \sum_{k=0}^{j-1}  \frac{1-\delta^{j-k}}{1-\delta} \nabla \mathcal{J}(\theta_k) \} \\
    & = \theta_0 - \alpha (1-\gamma) \sum_{k=0}^{t-1} \sum_{j=0}^{t-k-1} \gamma^{t-k-j-1} \frac{1-\delta^{j+1}}{1-\delta} \nabla \mathcal{J}(\theta_k) 
    \end{split}
\end{equation}

Difference of the coefficient for $\nabla \mathcal{J}(\theta_k)$ for each $k$ is 

\begin{equation}\label{eq:sup_pi_mt_delta}
    \begin{split}
        \frac{1-\delta^{t-k}}{1-\delta} - \sum_{j=0}^{t-k-1} (1-\gamma) \gamma^{t-k-j-1} \frac{1-\delta^{j+1}}{1-\delta} \geq 0
    \end{split}
\end{equation}

\begin{figure}[ht]
    \centering
    \includegraphics[width=.85\linewidth]{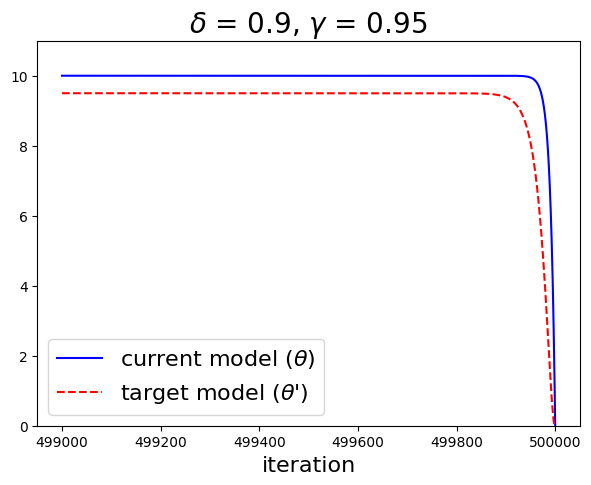}
    \caption{The difference of the first term (from current model $\theta$) and the second term (from target model $\theta'$) in \eqref{eq:sup_pi_mt_delta} for each iteration when $\delta$ is 0.9 and $\gamma$ is 0.95.
    The trend between iteration 0 to 499,000 is almost same with the early iteration of this figure. }
    \label{fig:sup_pi_vs_mt}
\end{figure}

Fig.\ref{fig:sup_pi_vs_mt} is the difference of the first term (from current model $\theta$) and the second term (from target model $\theta'$) in \eqref{eq:sup_pi_mt_delta} when $\delta$ is 0.9 and $\gamma$ is 0.95 same as our experiments.
We can see that the different between two terms is always greater or equal to 0.
Therefore, $\theta'$ is a more conservative target than $\theta$ in SGD with momentum optimizer, either.

\section{Experiment Settings}

\subsection{Dataset Details}
\label{sup:dataset}

We followed standard settings for CIFAR10 and SVHN.
For CIFAR10, there are 50,000 training images and 10,000 test images. We split the training set into a 45,000 train set and a 5,000 validation set for experiments.
The validation set consists of the same size per class.
We applied global contrast normalization and ZCA normalization.
For data augmentation, we used random horizontal flipping, random cropping by padding 2 pixels each side of the image, and added Gaussian noise with standard deviation 0.15 to each pixel.

For SVHN, there are 73,257 training images and 26,032 test images. We split the training set into a 65,931 train set and a 7,326 validation set for experiments.
The validation set consists of the same size per class.
We applied global contrast normalization and ZCA normalization.
For data augmentation, we used random cropping by padding 2 pixels on each side of the image only.

In our main experiments, we split the training set into the labeled set and the unlabeled set.
The size of the unlabeled set changes depending on unlabeled data imbalance types because of the limitation of the training dataset.
For fair experiments, we set the size of the unlabeled set based on the \textit{Same} case, which uses the lowest number of unlabeled samples.
The size of unlabeled data is described in Table.\ref{tab:cifar10_size_unlabel}, \ref{tab:svhn_size_unlabel}.

\begin{table}[ht]
    \centering
    \caption{Number of unlabeled data in CIFAR10 and SVHN according to imbalance factor and number of labeled data.}
    \label{tab:size_unlabel}
        \begin{subtable}{\linewidth}
        \centering
        \caption{CIFAR10}   
        \label{tab:cifar10_size_unlabel}
        \begin{tabular}{c|ccc}
        \toprule
         & \multicolumn{3}{c}{\# of labeled data} \\
         Imbalance Factor & 1k & 2k & 4k \\
         \midrule
         100 & 10166 & 9166 & 7166 \\
         50 & - & - & 8596 \\
         20 & - & - & 11322 \\
         10 & - & - & 14389 \\
         \bottomrule
        \end{tabular}
        \end{subtable}%
        \newline
        \vspace*{3mm}
        \begin{subtable}{\linewidth}
        \centering
        \caption{SVHN}   
        \label{tab:svhn_size_unlabel}
        \begin{tabular}{c|ccc}
        \toprule
         & \multicolumn{3}{c}{\# of labeled data} \\
         Imbalance Factor & 250 & 500 & 1k \\
         \midrule
         100 & 16109 & 15858 & 15360 \\
         50 & - & - & 17455 \\
         20 & - & - & 21449 \\
         10 & - & - & 25943 \\
         \bottomrule
        \end{tabular}
        \end{subtable}
\end{table}

\subsection{Implementation details}
\label{exp:detail}

In all experiments, we use the Wide-Resnet-28-2 model~\cite{zagoruyko2016wide}.
Following the settings from \citet{verma2019interpolation}, we set SGD with Nesterov momentum as our optimzer and adopted the cosine annealing technique~\cite{loshchilov2016sgdr}.
Detailed hyperparameters for experiments is described in Table.\ref{tab:hparams}.

\begin{table}[h]
    \caption{Hyperparameters for shared environment and each SSL algorithms and our method used in the experiments.}   
    \label{tab:hparams}
    \centering
    \adjustbox{max width=\linewidth}{
        \begin{tabular}{lc}
             \toprule 
             \multicolumn{2}{c}{Shared} \\
             \midrule 
             Training iteration & 500k \\
             Consistency ramp-up iteration & 200k \\ 
             Initial learning rate & 0.1 \\
             Cosine learning rate ramp-down iteration & 600k \\
             Weight decay & $10^{-4}$ \\
             Momentum & 0.9 \\
             \midrule
             \multicolumn{2}{c}{$\Pi$ Model} \\
             \midrule
             Max consistency coefficient & 20 \\
             \midrule
             \multicolumn{2}{c}{Mean Teacher} \\
             \midrule
             Max consistency coefficient & 8 \\
             Exponential Moving Average decay factor & 0.95 \\
             \midrule
             \multicolumn{2}{c}{VAT+em} \\
             \midrule
             Max consistency coefficient & 0.3 \\
             VAT $\epsilon$ (CIFAR10) & 6.0 \\
             VAT $\epsilon$ (SVHN) & 1.0\\
             VAT $\xi$ & $10^{e-6}$ \\
             \midrule
             \multicolumn{2}{c}{VAT$+$EM$+$SNTG (as for VAT)} \\
             \midrule
             Entropy penaly multiplier & 0.06 \\
             \midrule
             \multicolumn{2}{c}{Pseudo-Label} \\
             \midrule
             Max consistency coefficient & 1.0 \\
             Pseudo-label threshold & 0.95 \\
             \midrule
             \multicolumn{2}{c}{ICT} \\
             \midrule
             Max consistency coefficient & 100 \\
             Exponential Moving Average decay factor & 0.999 \\
             ICT $\alpha$ & 1.0 \\
             \midrule
             \multicolumn{2}{c}{Suppressed Consitency Loss (Ours)} \\
             \midrule
             Suppression Coefficient ($\beta$) & 0.5 \\
             \bottomrule
        \end{tabular}
    }
\end{table}

\section{Detailed Experiment Results}

We omitted the standard deviation from the experiment of the paper for readability.
Table.\ref{tab:sup_imbalance_factor},\ref{tab:sup_num_of_label},\ref{tab:sup_reweight} are tables with standard deviation.
Since we used five different seeds in each experiment, the class frequency distribution varies from seed to seed, which results in a change in baseline performance.
As a result, the standard deviation of our experiment is larger than that of the random initialization of the weights.

\begin{sidewaystable*}
    \centering
    \renewcommand{\tabcolsep}{0.7mm}
    \caption{Test error rates~(\%) and standard deviation from experiments with 4k number of labeled data and imbalance factor \{10, 20, 50, 100 \} under 3 different unlabeled imbalance types in CIFAR10 and SVHN.
    VAT+EM refers to Virtual Adversarial Training with Entropy Minimization.
    }
    \label{tab:sup_imbalance_factor}
    \begin{center}
    \scriptsize
    \adjustbox{max width=\linewidth}{
    \begin{subtable}{\linewidth}
    \centering
    \caption{CIFAR10}
    \label{tab:sup_cifar_imbalance_factor}
    \begin{tabularx}{.97\linewidth}{l||r|r|r|r||r|r|r|r||r|r|r|r}
        \toprule
        Unlabel Imbalance Type & \multicolumn{4}{c||}{Uniform ($\rho_u=1$)} & \multicolumn{4}{c||}{Half ($\rho_u = \rho_l/2$)} & \multicolumn{4}{c}{Same ($\rho_u=\rho_l$)} \\
        \midrule
        Imbalance factor ($\rho_l$) & \multicolumn{1}{c|}{10} & \multicolumn{1}{c|}{20} & \multicolumn{1}{c|}{50} & \multicolumn{1}{c||}{100} & \multicolumn{1}{c|}{10} & \multicolumn{1}{c|}{20} & \multicolumn{1}{c|}{50} & \multicolumn{1}{c||}{100} & \multicolumn{1}{c|}{10} & \multicolumn{1}{c|}{20} & \multicolumn{1}{c|}{50} & \multicolumn{1}{c}{100}  \\
        \midrule
Supervised & 23.03 $\pm$ 1.65 & 27.49 $\pm$ 1.87 & 33.15 $\pm$ 2.83 & 36.71 $\pm$ 2.79 & 23.03 $\pm$ 1.65 & 27.49 $\pm$ 1.87 & 33.15 $\pm$ 2.83 & 36.71 $\pm$ 2.79 & 23.03 $\pm$ 1.65 & 27.49 $\pm$ 1.87 & 33.15 $\pm$ 2.83 & 36.71 $\pm$ 2.79 \\
$\Pi$-Model \cite{laine2016temporal} & 21.1 $\pm$ 1.93 & 25.74 $\pm$ 3.82 & 33.91 $\pm$ 3.49 & 39.36 $\pm$ 4.47 & 22.69 $\pm$ 1.99 & 27.72 $\pm$ 4.17 & 33.96 $\pm$ 3.19 & 38.84 $\pm$ 4.17 & 23.49 $\pm$ 2.69 & 28.18 $\pm$ 3.31 & 34.22 $\pm$ 3.19 & 38.05 $\pm$ 3.19 \\
MT \cite{tarvainen2017mean} & 16.45 $\pm$ 1.24 & 19.25 $\pm$ 1.99 & 23.45 $\pm$ 3.30 & 29.06 $\pm$ 5.13 & 19.48 $\pm$ 1.96 & 23.30 $\pm$ 2.85 & 30.06 $\pm$ 3.92 & 35.37 $\pm$ 3.52 & 20.50 $\pm$ 2.58 & 24.67 $\pm$ 2.60 & 31.77 $\pm$ 3.79 & 35.91 $\pm$ 3.70 \\
VAT + em \cite{miyato2018virtual} & 17.93 $\pm$ 2.12 & 20.18 $\pm$ 3.18 & 30.43 $\pm$ 6.18 & 36.57 $\pm$ 7.20 & 20.17 $\pm$ 2.49 & 24.50 $\pm$ 2.88 & 32.54 $\pm$ 4.61 & 36.77 $\pm$ 3.75 & 21.45 $\pm$ 1.88 & 25.83 $\pm$ 3.21 & 33.13 $\pm$ 3.67 & 37.67 $\pm$ 2.20 \\
VAT + em + SNTG \cite{luo2018smooth} & 18.15 $\pm$ 2.25 & 20.39 $\pm$ 2.46 & 29.77 $\pm$ 6.71 & 36.34 $\pm$ 6.54 & 20.41 $\pm$ 2.47 & 24.64 $\pm$ 2.79 & 32.56 $\pm$ 4.05 & 38.48 $\pm$ 3.87 & 21.87 $\pm$ 2.65 & 26.49 $\pm$ 3.07 & 33.36 $\pm$ 3.86 & 38.48 $\pm$ 2.96 \\
Pseudo-Label \cite{lee2013pseudo} & 19.33 $\pm$ 1.36 & 24.34 $\pm$ 4.06 & 34.18 $\pm$ 4.23 & 39.59 $\pm$ 5.70 & 21.23 $\pm$ 2.52 & 26.78 $\pm$ 3.41 & 34.12 $\pm$ 4.51 & 39.72 $\pm$ 4.20 & 22.73 $\pm$ 2.74 & 27.50 $\pm$ 3.39 & 34.91 $\pm$ 2.57 & 38.69 $\pm$ 4.28 \\
ICT \cite{verma2019interpolation} & 18.01 $\pm$ 1.28 & 20.52 $\pm$ 1.91 & 30.18 $\pm$ 2.63 & 38.33 $\pm$ 4.72 & 19.53 $\pm$ 1.41 & 23.90 $\pm$ 2.07 & 31.09 $\pm$ 3.35 & 37.36 $\pm$ 2.02 & 19.96 $\pm$ 1.05 & 25.63 $\pm$ 1.91 & 33.56 $\pm$ 3.14 & 36.85 $\pm$ 3.44 \\
MT+SCL (ours) & 15.65 $\pm$ 0.69 & 16.99 $\pm$ 1.31 & 19.95 $\pm$ 2.36 & 22.62 $\pm$ 3.54 & 17.36 $\pm$ 1.17 & 21.74 $\pm$ 2.15 & 28.20 $\pm$ 3.09 & 33.09 $\pm$ 3.63 & 18.69 $\pm$ 2.09 & 22.98 $\pm$ 2.33 & 29.76 $\pm$ 2.40 & 34.22 $\pm$ 3.50 \\
         \bottomrule
    \end{tabularx}
    \end{subtable}%
    }
    \adjustbox{max width=\linewidth}{
    \begin{subtable}{\linewidth}
    \centering
    \bigskip
    \caption{SVHN}
    \label{tab:sup_svhn_imbalance_factor}
    \begin{tabularx}{.97\linewidth}{l||r|r|r|r||r|r|r|r||r|r|r|r}
        \toprule
        Unlabel Imbalance Type & \multicolumn{4}{c||}{Uniform ($\rho_u=1$)} & \multicolumn{4}{c||}{Half ($\rho_u = \rho_l/2$)} & \multicolumn{4}{c}{Same ($\rho_u=\rho_l$)} \\
        \midrule
        Imbalance factor ($\rho_l$) & \multicolumn{1}{c|}{10} & \multicolumn{1}{c|}{20} & \multicolumn{1}{c|}{50} & \multicolumn{1}{c||}{100} & \multicolumn{1}{c|}{10} & \multicolumn{1}{c|}{20} & \multicolumn{1}{c|}{50} & \multicolumn{1}{c||}{100} & \multicolumn{1}{c|}{10} & \multicolumn{1}{c|}{20} & \multicolumn{1}{c|}{50} & \multicolumn{1}{c}{100}  \\
        \midrule
Supervised & 18.49 $\pm$ 1.90 & 21.92 $\pm$ 2.28 & 30.03 $\pm$ 3.83 & 35.89 $\pm$ 6.39 & 18.49 $\pm$ 1.90 & 21.92 $\pm$ 2.28 & 30.03 $\pm$ 3.83 & 35.89 $\pm$ 6.39 & 18.49 $\pm$ 1.90 & 21.92 $\pm$ 2.28 & 30.03 $\pm$ 3.83 & 35.89 $\pm$ 6.39 \\
$\Pi$-Model \cite{laine2016temporal} & 11.74 $\pm$ 1.80 & 13.42 $\pm$ 2.14 & 21.63 $\pm$ 4.58 & 28.59 $\pm$ 7.90 & 12.96 $\pm$ 1.26 & 16.70 $\pm$ 4.01 & 24.02 $\pm$ 3.97 & 33.73 $\pm$ 7.52 & 13.46 $\pm$ 2.13 & 17.13 $\pm$ 2.61 & 26.53 $\pm$ 3.43 & 33.71 $\pm$ 8.17 \\
MT \cite{tarvainen2017mean} & 6.52 $\pm$ 0.55 & 6.75 $\pm$ 0.49 & 7.60 $\pm$ 1.85 & 8.94 $\pm$ 2.12 & 7.25 $\pm$ 0.38 & 8.85 $\pm$ 1.10 & 12.19 $\pm$ 1.68 & 17.23 $\pm$ 2.44 & 8.62 $\pm$ 1.29 & 9.29 $\pm$ 1.41 & 15.16 $\pm$ 3.54 & 21.01 $\pm$ 4.14 \\
VAT + em \cite{miyato2018virtual} & 6.81 $\pm$ 0.30 & 7.70 $\pm$ 0.87 & 13.84 $\pm$ 6.17 & 29.15 $\pm$ 4.80 & 8.99 $\pm$ 1.21 & 11.59 $\pm$ 1.85 & 18.95 $\pm$ 4.49 & 30.44 $\pm$ 6.95 & 10.39 $\pm$ 0.96 & 13.62 $\pm$ 2 & 21.49 $\pm$ 5.27 & 32.39 $\pm$ 8.25 \\
VAT + em + SNTG \cite{luo2018smooth} & 93.30 $\pm$ 0.00 & 93.30 $\pm$ 0.00 & 14.88 $\pm$ 5.38 & 93.30 $\pm$ 0.00 & 93.30 $\pm$ 0.00 & 93.30 $\pm$ 0.00 & 20.60 $\pm$ 5.73 & 93.30 $\pm$ 0.00 & 93.30 $\pm$ 0.00 & 93.30 $\pm$ 0.00 & 23.52 $\pm$ 7.34 & 93.30 $\pm$ 0.00 \\
Pseudo-Label \cite{lee2013pseudo} & 10.15 $\pm$ 0.87 & 9.97 $\pm$ 1.45 & 16.00 $\pm$ 4.34 & 32.79 $\pm$ 7.62 & 11.59 $\pm$ 1.96 & 13.97 $\pm$ 2.11 & 24.40 $\pm$ 4.46 & 33.70 $\pm$ 6.89 & 12.34 $\pm$ 1.79 & 15.93 $\pm$ 2.43 & 25.66 $\pm$ 5.95 & 33.53 $\pm$ 8.08 \\
ICT \cite{verma2019interpolation} & 27.82 $\pm$ 5.12 & 37.75 $\pm$ 7.50 & 58.20 $\pm$ 9.38 & 67.02 $\pm$ 12.66 & 22.38 $\pm$ 7.89 & 38.12 $\pm$ 6.57 & 48.88 $\pm$ 8.33 & 58.99 $\pm$ 7.35 & 24.53 $\pm$ 12.62 & 37.25 $\pm$ 8.22 & 49.85 $\pm$ 7.74 & 56.97 $\pm$ 10.28 \\
MT+SCL (ours) & 6.52 $\pm$ 0.53 & 7.11 $\pm$ 0.30 & 7.70 $\pm$ 0.73 & 8.56 $\pm$ 0.86 & 7.54 $\pm$ 0.50 & 9.29 $\pm$ 1.48 & 11.46 $\pm$ 1.21 & 18.63 $\pm$ 3.97 & 8.22 $\pm$ 0.89 & 10.04 $\pm$ 0.82 & 15.48 $\pm$ 2.29 & 20.39 $\pm$ 4.10 \\        
        \bottomrule
    \end{tabularx}
    \end{subtable}
    }
    \end{center}
\end{sidewaystable*}

\begin{sidewaystable*}
    \centering
    \renewcommand{\tabcolsep}{0.7mm}
    \caption{Test error rates~(\%) and standard deviation from experiments with imbalance factor 100 and the number of labeled data \{1k, 2k, 4k\} in CIFAR10, and the number of labeled data \{250, 500, 1k\} in SVHN under 3 different unlabeled imbalance types. 
    }
    \label{tab:sup_num_of_label}
    \begin{center}
    \footnotesize
    \adjustbox{max width=\linewidth}{
    \begin{subtable}{\linewidth}
    \centering
    \caption{CIFAR10}
    \label{tab:sup_cifar_num_of_label}
    \begin{tabularx}{.95\linewidth}{l||r|r|r||r|r|r||r|r|r}
        \toprule
        Unlabel Imbalance Type & \multicolumn{3}{c||}{Uniform ($\rho_u=1$)} & \multicolumn{3}{c||}{Half ($\rho_u = \rho_l/2$)} & \multicolumn{3}{c}{Same ($\rho_u=\rho_l$)} \\
        \midrule
         \# labeled data & \multicolumn{1}{c|}{1000} & \multicolumn{1}{c|}{2000} & \multicolumn{1}{c||}{4000} & \multicolumn{1}{c|}{1000} & \multicolumn{1}{c|}{2000} & \multicolumn{1}{c||}{4000} & \multicolumn{1}{c|}{1000} & \multicolumn{1}{c|}{2000} & \multicolumn{1}{c}{4000}\\
        \midrule
Supervised & 54.24 $\pm$ 2.08 & 45.81 $\pm$ 3.00 & 36.71 $\pm$ 2.79 & 54.24 $\pm$ 2.08 & 45.81 $\pm$ 3 & 36.71 $\pm$ 2.79 & 54.24 $\pm$ 2.08 & 45.81 $\pm$ 3 & 36.71 $\pm$ 2.79 \\
$\Pi$-Model \cite{laine2016temporal} & 56.82 $\pm$ 3.63 & 48.55 $\pm$ 4.26 & 39.36 $\pm$ 4.47 & 55.99 $\pm$ 2.79 & 47.74 $\pm$ 3.82 & 38.84 $\pm$ 4.17 & 55.42 $\pm$ 1.47 & 46.83 $\pm$ 3.29 & 38.05 $\pm$ 3.19 \\
MT \cite{tarvainen2017mean} & 51.74 $\pm$ 5.33 & 38.94 $\pm$ 7.67 & 29.06 $\pm$ 5.13 & 51.61 $\pm$ 4.58 & 42.47 $\pm$ 5.66 & 35.37 $\pm$ 3.52 & 52.58 $\pm$ 3.23 & 44.11 $\pm$ 4.16 & 35.91 $\pm$ 3.7 \\
VAT + em \cite{miyato2018virtual} & 53.68 $\pm$ 4.21 & 48.47 $\pm$ 3.66 & 36.57 $\pm$ 7.20 & 53.60 $\pm$ 3.18 & 45.20 $\pm$ 4.84 & 36.77 $\pm$ 3.75 & 53.62 $\pm$ 3.14 & 44.77 $\pm$ 2.82 & 37.67 $\pm$ 2.2 \\
VAT + em + SNTG \cite{luo2018smooth} & 54.53 $\pm$ 3.09 & 48.23 $\pm$ 3.50 & 36.34 $\pm$ 6.54 & 55.59 $\pm$ 3.54 & 45.37 $\pm$ 3.25 & 38.48 $\pm$ 3.87 & 55.55 $\pm$ 2.47 & 45.99 $\pm$ 4.29 & 38.48 $\pm$ 2.96 \\
Pseudo-Label \cite{lee2013pseudo} & 58.19 $\pm$ 1.73 & 50.01 $\pm$ 2.78 & 39.59 $\pm$ 5.70 & 57.05 $\pm$ 2.86 & 49.42 $\pm$ 2.42 & 39.72 $\pm$ 4.20 & 56.68 $\pm$ 3.00 & 48.45 $\pm$ 3 & 38.69 $\pm$ 4.28 \\
ICT \cite{verma2019interpolation} & 57.10 $\pm$ 4.56 & 48.25 $\pm$ 1.53 & 38.33 $\pm$ 4.72 & 56.02 $\pm$ 3.37 & 47.60 $\pm$ 2.39 & 37.36 $\pm$ 2.02 & 55.10 $\pm$ 2.68 & 47.19 $\pm$ 1.57 & 36.85 $\pm$ 3.44 \\
MT+SCL (ours) & 42.84 $\pm$ 2.88 & 28.69 $\pm$ 4.55 & 22.62 $\pm$ 3.54 & 45.72 $\pm$ 2.62 & 39.97 $\pm$ 2.58 & 33.09 $\pm$ 3.63 & 48.00 $\pm$ 3.41 & 40.69 $\pm$ 3.41 & 34.22 $\pm$ 3.50 \\
         \bottomrule
    \end{tabularx}
    \end{subtable}%
    }
    \adjustbox{max width=.95\linewidth}{
    \begin{subtable}{\linewidth}
    \centering
    \bigskip
    \caption{SVHN}
    \label{tab:sup_svhn_num_of_label}
    \begin{tabularx}{\linewidth}{l||r|r|r||r|r|r||r|r|r}
        \toprule
        Unlabel Imbalance Type & \multicolumn{3}{c||}{Uniform ($\rho_u=1$)} & \multicolumn{3}{c||}{Half ($\rho_u = \rho_l/2$)} & \multicolumn{3}{c}{Same ($\rho_u=\rho_l$)} \\
        \midrule
         \# labeled data & \multicolumn{1}{c|}{1000} & \multicolumn{1}{c|}{2000} & \multicolumn{1}{c||}{4000} & \multicolumn{1}{c|}{1000} & \multicolumn{1}{c|}{2000} & \multicolumn{1}{c||}{4000} & \multicolumn{1}{c|}{1000} & \multicolumn{1}{c|}{2000} & \multicolumn{1}{c}{4000}\\
        \midrule
Supervised & 61.31 $\pm$ 8.05 & 47.98 $\pm$ 7.08 & 35.89 $\pm$ 6.39 & 61.31 $\pm$ 8.05 & 47.98 $\pm$ 7.08 & 35.89 $\pm$ 6.39 & 61.31 $\pm$ 8.05 & 47.98 $\pm$ 7.08 & 35.89 $\pm$ 6.39 \\
$\Pi$-Model \cite{laine2016temporal} & 54.51 $\pm$ 8.43 & 39.49 $\pm$ 9.10 & 28.59 $\pm$ 7.90 & 54.14 $\pm$ 9.11 & 42.20 $\pm$ 7.73 & 33.73 $\pm$ 7.52 & 54.10 $\pm$ 10.07 & 43.89 $\pm$ 9.68 & 33.71 $\pm$ 8.17 \\
MT \cite{tarvainen2017mean} & 38.32 $\pm$ 11.69 & 18.14 $\pm$ 10.47 & 8.94 $\pm$ 2.12 & 41.72 $\pm$ 9.34 & 23.33 $\pm$ 10.78 & 17.23 $\pm$ 2.44 & 42.42 $\pm$ 9.74 & 28.86 $\pm$ 10.57 & 21.01 $\pm$ 4.14 \\
VAT + em \cite{miyato2018virtual} & 64.67 $\pm$ 6.41 & 44.04 $\pm$ 8.88 & 29.15 $\pm$ 4.80 & 58.01 $\pm$ 10.44 & 41.15 $\pm$ 10.23 & 30.44 $\pm$ 6.95 & 55.03 $\pm$ 8.85 & 42.44 $\pm$ 8.04 & 32.39 $\pm$ 8.25 \\
VAT + em + SNTG \cite{luo2018smooth} & 65.02 $\pm$ 5.23 & 93.30 $\pm$ 0.00 & 93.30 $\pm$ 0.00 & 57.94 $\pm$ 8.87 & 93.30 $\pm$ 0.00 & 93.30 $\pm$ 0.00 & 54.19 $\pm$ 9.43 & 93.30 $\pm$ 0.00 & 93.30 $\pm$ 0.00 \\
Pseudo-Label \cite{lee2013pseudo} & 63.16 $\pm$ 6.60 & 49.78 $\pm$ 7.92 & 32.79 $\pm$ 7.62 & 54.79 $\pm$ 10.42 & 44.32 $\pm$ 7.29 & 33.70 $\pm$ 6.89 & 56.83 $\pm$ 8.81 & 43.71 $\pm$ 5.76 & 33.53 $\pm$ 8.08 \\
ICT \cite{verma2019interpolation} & 86.54 $\pm$ 5.27 & 77.64 $\pm$ 1.94 & 67.02 $\pm$ 12.66 & 84.22 $\pm$ 7.46 & 72.21 $\pm$ 9.43 & 58.99 $\pm$ 7.35 & 85.15 $\pm$ 5.89 & 71.19 $\pm$ 7.68 & 56.97 $\pm$ 10.28 \\
MT+SCL (ours) & 26.25 $\pm$ 12.84 & 15.31 $\pm$ 6.81 & 8.56 $\pm$ 0.86 & 33.44 $\pm$ 10.81 & 22.26 $\pm$ 6.22 & 18.63 $\pm$ 3.97 & 35.32 $\pm$ 10.59 & 27.13 $\pm$ 10.58 & 20.39 $\pm$ 4.10 \\
        \bottomrule
    \end{tabularx}
    \end{subtable}
    }
    \end{center}
\end{sidewaystable*}

\begin{sidewaystable*}
    \centering
    \renewcommand{\tabcolsep}{0.7mm}
    \caption{Test error rates~(\%) and standard deviation from experiments with different re-weighting methods in CIFAR10 and SVHN. We compared inverse and normalization (IN), focal loss (FOCAL), and class-balanced loss (CB) to conventional cross-entropy loss (CE).
    }
    \label{tab:sup_reweight}
    \begin{center}
    \scriptsize
    \adjustbox{max width=\linewidth}{
    \begin{subtable}{\linewidth}
    \centering
    \caption{CIFAR10}
    \label{tab:sup_cifar_reweight}
    \begin{tabularx}{.95\linewidth}{l||r|r|r|r||r|r|r|r||r|r|r|r}
        \toprule
        Unlabel Imbalance Type & \multicolumn{4}{c||}{Uniform ($\rho_u=1$)} & \multicolumn{4}{c||}{Half ($\rho_u = \rho_l/2$)} & \multicolumn{4}{c}{Same ($\rho_u=\rho_l$)} \\
         Re-weighting Method & \multicolumn{1}{c|}{CE} & \multicolumn{1}{c|}{IN} & \multicolumn{1}{c|}{Focal} & \multicolumn{1}{c||}{CB} & \multicolumn{1}{c|}{CE} & \multicolumn{1}{c|}{IN} & \multicolumn{1}{c|}{Focal} & \multicolumn{1}{c||}{CB} & \multicolumn{1}{c|}{CE} & \multicolumn{1}{c|}{IN} & \multicolumn{1}{c|}{Focal} & \multicolumn{1}{c}{CB} \\
         \midrule
Supervised & 36.71 $\pm$ 2.79 & 35.73 $\pm$ 2.39 & 36.8 $\pm$ 2.41 & 37.19 $\pm$ 2.88 & 36.71 $\pm$ 2.79 & 35.73 $\pm$ 2.39 & 36.80 $\pm$ 2.41 & 37.19 $\pm$ 2.88 & 36.71 $\pm$ 2.79 & 35.73 $\pm$ 2.39 & 36.8 $\pm$ 2.41 & 37.19 $\pm$ 2.88 \\
$\Pi$-Model \cite{laine2016temporal} & 39.36 $\pm$ 4.47 & 36.90 $\pm$ 3.56 & 39.89 $\pm$ 4.23 & 39.20 $\pm$ 4.38 & 38.84 $\pm$ 4.17 & 37.82 $\pm$ 1.55 & 38.28 $\pm$ 3.37 & 37.51 $\pm$ 1.43 & 38.05 $\pm$ 3.19 & 37.18 $\pm$ 2.12 & 38.1 $\pm$ 3.37 & 37.34 $\pm$ 2.48 \\
MT \cite{tarvainen2017mean} & 29.06 $\pm$ 5.13 & 24 $\pm$ 3.17 & 30.73 $\pm$ 6.2 & 29.5 $\pm$ 5.69 & 35.37 $\pm$ 3.52 & 33.08 $\pm$ 2.78 & 34.45 $\pm$ 3.89 & 35.04 $\pm$ 3.42 & 35.91 $\pm$ 3.70 & 34.01 $\pm$ 2.85 & 35.65 $\pm$ 2.64 & 35.17 $\pm$ 3.77 \\
VAT + em \cite{miyato2018virtual} & 36.57 $\pm$ 7.20 & 31.34 $\pm$ 5.01 & 37.51 $\pm$ 7.56 & 36.78 $\pm$ 8.40 & 36.77 $\pm$ 3.75 & 36.20 $\pm$ 1.93 & 37.62 $\pm$ 3.94 & 38.13 $\pm$ 4.63 & 37.67 $\pm$ 2.20 & 36.91 $\pm$ 2.33 & 37.88 $\pm$ 3.66 & 37.64 $\pm$ 2.74 \\
VAT + em + SNTG \cite{luo2018smooth} & 36.34 $\pm$ 6.54 & 33.03 $\pm$ 4.78 & 37.78 $\pm$ 6.94 & 36.26 $\pm$ 6.78 & 38.48 $\pm$ 3.87 & 35.90 $\pm$ 2.78 & 38.01 $\pm$ 4.85 & 37.44 $\pm$ 4 & 38.48 $\pm$ 2.96 & 36.99 $\pm$ 2.77 & 37.71 $\pm$ 4.32 & 37.53 $\pm$ 3.52 \\
Pseudo-Label \cite{lee2013pseudo} & 39.59 $\pm$ 5.70 & 30.62 $\pm$ 3.62 & 37.90 $\pm$ 6.87 & 39.38 $\pm$ 4.79 & 39.72 $\pm$ 4.20 & 37.36 $\pm$ 3.14 & 38.77 $\pm$ 4.23 & 39.12 $\pm$ 3.19 & 38.69 $\pm$ 4.28 & 36.84 $\pm$ 3.31 & 38.92 $\pm$ 3.31 & 38.52 $\pm$ 3.13 \\
MT+SCL (ours) & 22.62 $\pm$ 3.54 & 21.59 $\pm$ 3.05 & 23.44 $\pm$ 3.24 & 22.93 $\pm$ 3.53 & 33.09 $\pm$ 3.63 & 31.63 $\pm$ 2.31 & 34.09 $\pm$ 3.22 & 33.14 $\pm$ 3.43 & 34.22 $\pm$ 3.50 & 32.09 $\pm$ 2.16 & 33.93 $\pm$ 3.27 & 34.66 $\pm$ 4.37 \\
         \bottomrule
    \end{tabularx}
    \end{subtable}%
    }
    \adjustbox{max width=\linewidth}{
    \begin{subtable}{\linewidth}
    \centering
    \bigskip
    \caption{SVHN}
    \label{tab:sup_svhn_reweight}
    \begin{tabularx}{.95\linewidth}{l||r|r|r|r||r|r|r|r||r|r|r|r}
        \toprule
        Unlabel Imbalance Type & \multicolumn{4}{c||}{Uniform ($\rho_u=1$)} & \multicolumn{4}{c||}{Half ($\rho_u = \rho_l/2$)} & \multicolumn{4}{c}{Same ($\rho_u=\rho_l$)} \\
         Re-weighting Method & \multicolumn{1}{c|}{CE} & \multicolumn{1}{c|}{IN} & \multicolumn{1}{c|}{Focal} & \multicolumn{1}{c||}{CB} & \multicolumn{1}{c|}{CE} & \multicolumn{1}{c|}{IN} & \multicolumn{1}{c|}{Focal} & \multicolumn{1}{c||}{CB} & \multicolumn{1}{c|}{CE} & \multicolumn{1}{c|}{IN} & \multicolumn{1}{c|}{Focal} & \multicolumn{1}{c}{CB} \\
         \midrule
Supervised & 35.89 $\pm$ 6.39 & 34.60 $\pm$ 6.51 & 35.45 $\pm$ 6.20 & 35.3 $\pm$ 7.08 & 35.89 $\pm$ 6.39 & 34.60 $\pm$ 6.51 & 35.45 $\pm$ 6.20 & 35.30 $\pm$ 7.08 & 35.89 $\pm$ 6.39 & 34.60 $\pm$ 6.51 & 35.45 $\pm$ 6.20 & 35.30 $\pm$ 7.08 \\
$\Pi$-Model \cite{laine2016temporal} & 28.59 $\pm$ 7.90 & 26.72 $\pm$ 6.12 & 30.15 $\pm$ 8.13 & 27.99 $\pm$ 5.95 & 33.73 $\pm$ 7.52 & 29.60 $\pm$ 7.33 & 31.67 $\pm$ 6.43 & 31.12 $\pm$ 7.72 & 33.71 $\pm$ 8.17 & 31.7 $\pm$ 6.94 & 31.70 $\pm$ 5.08 & 33.17 $\pm$ 5.35 \\
MT \cite{tarvainen2017mean} & 8.94 $\pm$ 2.12 & 6.82 $\pm$ 0.34 & 8.66 $\pm$ 2.11 & 7.86 $\pm$ 1.82 & 17.23 $\pm$ 2.44 & 17.02 $\pm$ 3.93 & 16.20 $\pm$ 3.70 & 16.68 $\pm$ 2.24 & 21.01 $\pm$ 4.14 & 20.80 $\pm$ 5.43 & 20.01 $\pm$ 4.41 & 21.77 $\pm$ 4.45 \\
VAT + em \cite{miyato2018virtual} & 29.15 $\pm$ 4.80 & 20.26 $\pm$ 7.97 & 28.09 $\pm$ 5.46 & 29.37 $\pm$ 4.76 & 30.44 $\pm$ 6.95 & 27.44 $\pm$ 7.63 & 28.62 $\pm$ 8.11 & 29.65 $\pm$ 6.90 & 32.39 $\pm$ 8.25 & 29.18 $\pm$ 7.45 & 30.62 $\pm$ 8.23 & 30.93 $\pm$ 6.51 \\
VAT + em + SNTG \cite{luo2018smooth} & 93.30 $\pm$ 0.00 & 93.30 $\pm$ 0.00 & 93.30 $\pm$ 0.00 & 93.30 $\pm$ 0.00 & 93.30 $\pm$ 0.00 & 93.30 $\pm$ 0.00 & 93.30 $\pm$ 0.00 & 93.30 $\pm$ 0.00 & 93.30 $\pm$ 0.00 & 93.30 $\pm$ 0.00 & 93.30 $\pm$ 0.00 & 93.30 $\pm$ 0.00 \\
Pseudo-Label \cite{lee2013pseudo} & 32.79 $\pm$ 7.62 & 13.48 $\pm$ 2.45 & 35.07 $\pm$ 10.85 & 34.38 $\pm$ 9.48 & 33.70 $\pm$ 6.89 & 31.83 $\pm$ 4.98 & 32.79 $\pm$ 6.72 & 32.83 $\pm$ 8.27 & 33.53 $\pm$ 8.08 & 31.62 $\pm$ 6.06 & 33.63 $\pm$ 6.24 & 34.55 $\pm$ 6.58 \\
MT+SCL (ours) & 8.56 $\pm$ 0.86 & 8.48 $\pm$ 1.47 & 7.74 $\pm$ 0.58 & 9.02 $\pm$ 1.34 & 18.63 $\pm$ 3.97 & 18.59 $\pm$ 3.71 & 16.34 $\pm$ 2.62 & 16.44 $\pm$ 2.47 & 20.39 $\pm$ 4.10 & 20.51 $\pm$ 5.43 & 20.95 $\pm$ 4.48 & 21.06 $\pm$ 4.78 \\
         \bottomrule
    \end{tabularx}
    \end{subtable}
    }
    \end{center}
\end{sidewaystable*}

\section{Object Detection Experiment Settings}

\subsection{Dataset Details}
\label{od:dataset}

\begin{figure}[!t]
    \centering
    \begin{subfigure}{0.9\linewidth}
        \centering
        \includegraphics[width=8cm, height=7cm]{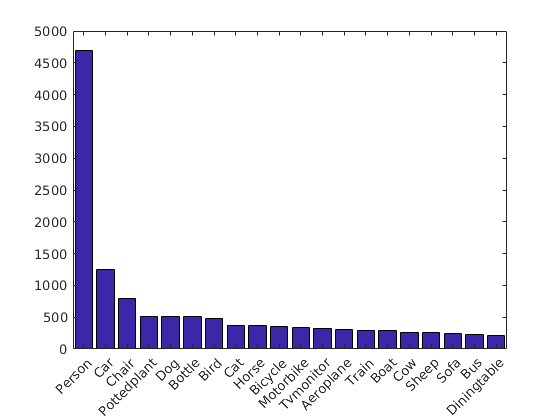}
        \caption{Labeled dataset (VOC2007)}
        \label{fig:voc07}
    \end{subfigure}
    \\
    \begin{subfigure}{0.9\linewidth}
        \centering
        \includegraphics[width=8cm, height=7cm]{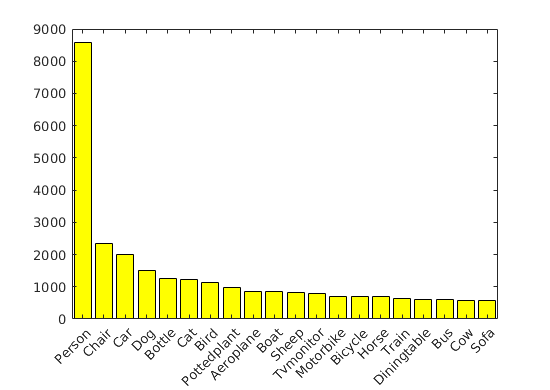}
        \caption{Unlabeled dataset (VOC2012)}
        \label{fig:voc12}
    \end{subfigure}
    
\vspace{-2mm}
\caption{Distributions for the labeled dataset (VOC2007) and the unlabeled dataset (VOC2012).}
\label{fig:voc}
\vspace{-2mm}
\end{figure}

We used PASCAL VOC2007 trainval dataset as the labeled data and PASCAL VOC2012 trainval dataset as the unlabeled data.
Fig.~\ref{fig:voc} shows the distributions of PASCAL VOC data.
The imbalance factor of labeled data is 22, and the imbalance factor of unlabeled data is 15.
The order of the number of classes is also different.
It means that the object detection task is more difficult and real settings.

\subsection{Implementation details}
\label{od:imple}

We followed the CSD\footnote{https://github.com/soo89/CSD-SSD} \cite{jeong2019consistency} experiment settings and used the SSD300 model \cite{liu2016ssd}. All hyperparameters such as coefficient, learning iteration, schedule function, and background elimination are the same. We set the $g(z)$ as $z / N_{max} $ becuase it shows better performance.

\end{document}